\documentclass[pdflatex, sn-basic, Numbered]{sn-jnl}

\usepackage{lmodern}
\usepackage[T1]{fontenc}
\usepackage{textcomp}

\usepackage{graphicx}%
\usepackage{multirow}%
\usepackage{amsmath,amssymb,amsfonts}%
\usepackage{amsthm}%
\usepackage{mathrsfs}%
\usepackage[title]{appendix}%
\usepackage{xcolor}%
\usepackage{textcomp}%
\usepackage{manyfoot}%
\usepackage{booktabs}%
\usepackage{algorithm}%
\usepackage{algorithmicx}%
\usepackage{algpseudocode}%
\usepackage{comment}
\usepackage{listings}%
\usepackage{subcaption}
\usepackage{enumitem}

\usepackage{hyperref}
\usepackage{eso-pic}

\DeclareMathOperator*{\argmin}{arg\,min}

\newcommand{\btheta}{\boldsymbol{\theta}}
\newcommand{\bxi}{\boldsymbol{\xi}}

\newcommand{\bx}{\mathbf{x}}

\theoremstyle{thmstyleone}%
\theoremstyle{thmstyletwo}%

\theoremstyle{thmstylethree}%

\begin{document}

\title{ASBI: Leveraging Informative Real-World Data for Active Black-Box Simulator Tuning}

\author*[1]{\fnm{Gahee} \sur{Kim}}\email{kim.gahee.kc3@is.naist.jp}
\author[1]{\fnm{Takamitsu} \sur{Matsubara}}\email{takam-m@is.naist.jp}
\affil[1]{\orgdiv{Graduate School of Science and Technology}, \orgname{Nara Institute of Science and Technology}, \orgaddress{\state{Nara}, \country{Japan}}}

\abstract{Black-box simulators are widely used in robotics, but optimizing their parameters remains challenging due to inaccessible likelihoods. Simulation-Based Inference (SBI) tackles this issue using simulation-driven approaches, estimating the posterior from offline real observations and forward simulations.
However, in black-box scenarios, preparing observations that contain sufficient information for parameter estimation is difficult due to the unknown relationship between parameters and observations. In this work, we present Active Simulation-Based Inference (ASBI), a parameter estimation framework that uses robots to actively collect real-world online data to achieve accurate black-box simulator tuning. Our framework optimizes robot actions to collect informative observations by maximizing information gain, which is defined as the expected reduction in Shannon entropy between the posterior and the prior. While calculating information gain requires the likelihood, which is inaccessible in black-box simulators, our method solves this problem by leveraging Neural Posterior Estimation (NPE), which leverages a neural network to learn the posterior estimator. 
Three simulation experiments quantitatively verify that our method achieves accurate parameter estimation, with posteriors sharply concentrated around the true parameters. Moreover, we show a practical application using a real robot to estimate the simulation parameters of cubic particles corresponding to two real objects, beads and gravel, with a bucket pouring action. }


\begingroup
  \renewcommand\thefootnote{}
  \footnotetext{\footnotesize
  This is the \textbf{preprint} version of an article accepted for publication in \textit{Applied Intelligence} (Springer, 2025). The final authenticated version will be available via \url{https://doi.org/10.1007/s10489-025-06934-z}.}
  \addtocounter{footnote}{-1}
\endgroup

\maketitle

\section{Introduction}\label{sec:sec1}

Modern simulators offer sophisticated dynamic models as black-box functions, enabling users to utilize them without understanding the underlying models. This ease of use has led to the widespread adoption of black-box simulators in a variety of fields.  However, the main obstacle to implementing black-box simulators is the need to optimize simulation parameters so that they align with real-world conditions. Since the likelihood of a given real-world observation (i.e., the function measuring how well the simulated observation matches real-world observations) is inaccessible in black-box simulators, it becomes challenging to infer the parameters.

To address this issue, Simulation-Based Inference (SBI)~\cite{Cranmer2020-cc} has emerged as a promising avenue, providing a simulation-driven approach to optimize simulation parameters without requiring the likelihood. SBI provides likelihood-free Bayesian inference by leveraging extensive forward simulations to generate a training dataset that can be used to infer the posterior of a given real observation without an explicit likelihood. This method has been successfully applied across various fields, including physics~\cite{Pina2020-xv}, neuroscience~\cite {Lueckmann2017-sa}, and astronomy~\cite{Vasist2023-cv, Dax2021-rt, Khullar2022-kv}.

\begin{figure}[tb]
\centering 
\includegraphics[width=0.7\hsize]{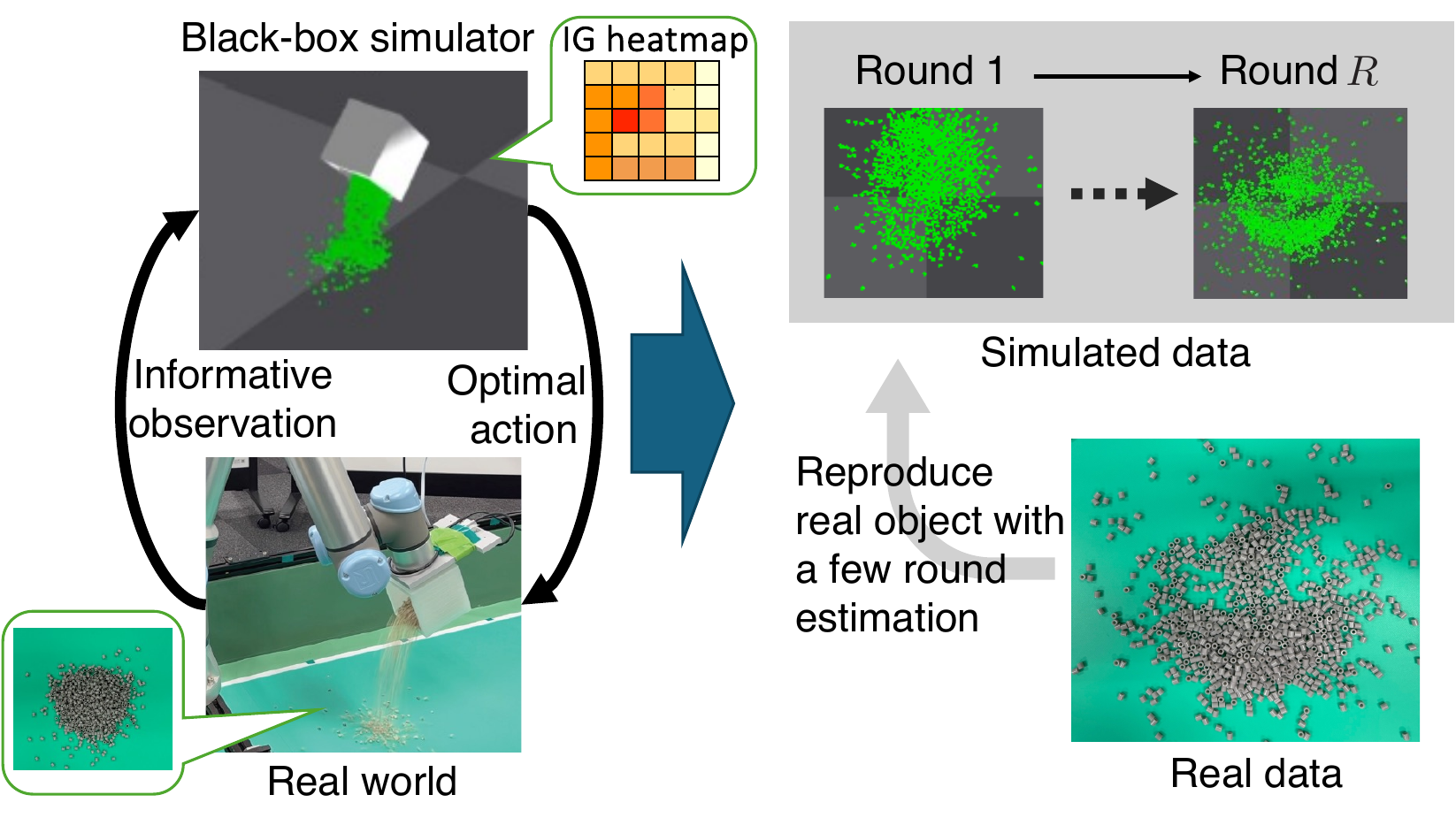}
\caption{Particle-parameter estimation task with ASBI. A robot collects real-world informative data and then optimizes the simulation parameters to match the simulated data to the real data. By selecting the optimal action that maximizes Information Gain  (IG) and gathering real-world observations online in each round, ASBI can achieve accurate parameter estimation with a small number of real executions.}
\label{fig:fig1}
\end{figure}

The aim of this study is to develop a practical parameter-tuning framework for black-box robotics simulators. In robotics, simulators are essential tools due to the high costs and safety risks associated with real-world experiments. Furthermore, as robotics technology advances, there is a growing demand for handling complicated materials such as cloth~\cite{Lips2024-bc}, ropes~\cite{Sundaresan2020-cv}, cables~\cite{Mou2022-py}, powders~\cite{Kadokawa2023-lc}, and soils~\cite{Egli2022-ca}, which are difficult to represent with traditional rigid-body simulators. Since these materials require expert-level knowledge of material dynamics, and building simulations from scratch is a challenging task, black-box simulators equipped with these advanced models as black-box features (e.g.,~\cite{Mittal2023-yd, algoryx-2025}) are now actively used in robotics. While these simulators make experiments easier, users need to tune the simulation parameters to accurately reflect real-world conditions. Consequently, developing a parameter estimation method that can automatically optimize the parameters of black-box simulators has become an urgent task. 

However, previous work on SBI is not well-suited to black-box simulators in robotics. SBI assumes that a given offline real-world observation contains sufficient information about all parameters. Yet in black-box simulators, it is difficult to determine whether an observation actually contains useful information about the parameters. This problem can be solved by collecting offline data across all possible actions and combining the resulting posteriors. However, in robotics, where real executions are expensive, the amount of available real observation data is limited, making the estimated posterior highly uncertain. 

To address the above issues, we introduce an online data collection approach for SBI instead of relying on offline data.  By interacting with the real world through exploratory robot actions, robots may collect additional data to effectively reduce prediction uncertainty. If robots can strategically collect informative real observations, this approach can solve the limitations associated with offline data collection. 

Specifically, we propose Active Simulation-Based Inference (ASBI), a framework that facilitates efficient and accurate parameter estimation in black-box simulators by having a robot actively interact with the real world and collect observations online. ASBI offers a key advantage in its ability to adaptively select optimal actions based on the current belief (prior), allowing the robot to collect informative observations from the environment in each round. Compared to conventional SBI methods that rely on static, offline datasets, ASBI dynamically refines the posterior using observations collected through actions specifically chosen to maximize the expected reduction in uncertainty. This sequential adaptation facilitates accurate parameter estimation with only a small number of data collections. Such sample efficiency is particularly beneficial in robotics, where real-world executions are often costly and time-consuming.

To select the optimal action that produces an accurate estimate, ASBI evaluates actions based on information gain~\cite{Lindley1956-ge}, which measures the expected reduction in uncertainty between the posterior and the prior. ASBI consists of two main components:

\begin{enumerate}
\item{\bf Likelihood-free posterior estimator with action variables:} To handle robot actions as variables, ASBI straightforwardly extends the structure of Neural Posterior Estimator (NPE)~\cite{Papamakarios2016-ki, Lueckmann2017-sa, Greenberg2019-on}, a state-of-the-art SBI method, to include actions as variables. This action-extended NPE estimates a posterior conditioned on both observations and actions.

\item{\bf Likelihood-free action optimizer via information gain:} While the analytical calculation of information gain requires the likelihood, ASBI approximates it using the action-extended NPE and forward simulations. This simulation-driven approach enables ASBI to select the optimal action without the likelihood.
\end{enumerate}

To verify the effectiveness of ASBI, we first evaluate it on a numerical toy model using three comparison methods: (1) naive sequential parameter estimation, which selects a robot action randomly at each round (NSBI); (2) likelihood-based active parameter estimation (ALHI), which uses surrogate likelihood models to calculate information gain; and (3) Sequential Bayesian Experimental Design (SeqBED)~\cite{Kleinegesse2021-zj}, which is the most relevant method, primarily applied in biological modeling. We restrict the comparison with SeqBED to the toy model, as it requires a prohibitively large number of forward simulations, making it impractical for robotics applications. We then apply ASBI to parameter estimation tasks using actual black-box robotics simulators capable of modeling complex physical interactions. We evaluate two tasks: a {\bf box-collision task}, in which mass and friction parameters are estimated from object collision dynamics, and a {\bf particle-parameter estimation task}, where cubic particles are poured using a bucket mounted on a robot manipulator (Fig.~\ref{fig:fig1}). For both tasks, we conduct sim-to-sim experiments to quantitatively evaluate ASBI using robotics simulators. Additionally, we perform a real-to-sim experiment for the particle-parameter estimation task, in which we estimate the simulation parameters corresponding to two real-world materials, beads and gravel, that exhibit different physical properties.

\section{Related Work}\label{sec:sec2}

\subsection{Simulation-based Inference (SBI)}

SBI is a collective term for approaches that use simulation data as a training dataset to learn the posterior when the exact likelihood is inaccessible. Since SBI enables inference without likelihoods, it is also called likelihood-free inference. Approximation Bayesian Computation (ABC)~\cite{Pritchard1999-th, Beaumont2002-sj, Marjoram2003-th, Bonassi2015-ly} is a traditional SBI method that approximates the posterior by accepting parameters that produce simulated data within a certain distance from the given observation data. However, ABC approaches do not scale well to high-dimensional data, since the required number of samples grows rapidly with the dimension of the parameter space. Neural Posterior Estimation (NPE)~\cite{Papamakarios2016-ki} provides an amortized inference method by training a neural network to approximate the posterior. For a given observation, Sequential Neural Posterior Estimation (SNPE)~\cite{Papamakarios2016-ki, Lueckmann2017-sa, Greenberg2019-on} enhances sampling efficiency by dividing neural network training into multiple rounds and gradually changing the sampling distribution of the training dataset.

Compared to such previous works, our method considers online data collection using robots in the real world. In black-box simulators where the relationship between parameters and observations is unknown, preparing offline data that are informative for all parameters becomes infeasible as the model complexity increases. We train a neural network through multiple rounds similar to SNPE, but in each round, we strategically select the robot action to gather an informative real observation. This online data collection with action selection facilitates better performance by combining informative real observations. Furthermore, our approach differs from online BayesSim~\cite{Possas2020-yh} and Neural Posterior Domain Randomization~\cite{Muratore2022-cy}, which apply SNPE to robotics and collect online real data, in that they do not consider strategic selection of robot actions to improve estimation accuracy. 

\subsection{Likelihood-based active estimation}

Parameter estimation with adaptively selected actions has proven its effectiveness in various fields, including psychology~\cite{Myung2013-wh}, physics~\cite{Dushenko2020-xd}, and biology~\cite{Kleinegesse2021-zj} (see ~\cite{Rainforth2024-kn, Ryan2016-mo} for a comprehensive review). Information gain~\cite{Lindley1956-ge} is a commonly used criterion measuring the expected changes in Shannon entropy from prior to posterior. To evaluate information gain in black-box scenarios, previous works~\cite{Saal2010-uf, Cooper2021-qg, Dutta2023-qq} train surrogate likelihood models and analytically compute the posterior using the learned likelihood. 

While this surrogate likelihood model allows the analytic calculation of information gain for black-box simulators, training likelihood in robotics simulation struggles with modeling errors because the output in robotics is often high-dimensional, such as images and point clouds. The proposed method mitigates this problem by adopting a likelihood-free approach to directly training the posterior on parameter spaces, which is, in general significantly lower-dimensional than observation spaces. 

\subsection{Likelihood-free active estimation}

Recent studies \cite{Margolis2023-ts, Memmel2024-kk} have employed reinforcement learning (RL) to learn robot policies for improving estimation accuracy. These approaches differ from our method in that they optimize robot policy in an offline manner and acquire a universal policy that can be deployed in any real environment. On the other hand, our approach focuses on the scenario where the real environment we estimate is already determined. Instead of using offline policy training, ASBI selects the optimal action in an online manner for efficient use of simulation resources. 

Outside the field of robotics, Sequential Bayesian Experimental Design (SeqBED)~\cite{Kleinegesse2021-zj} uses likelihood-to-marginal ratio estimation \cite{Thomas2022-eu} to maximize information gain. This method avoids explicit likelihood modeling, but it performs independent binary classification on the observation space for each parameter candidate, determining whether an observation came from the likelihood or the marginal. This method requires significant computation resources, which is impractical in online estimation for robotics simulators. Our approach overcomes these limitations through amortized training of a neural network. 

\vspace{10pt}
As reviewed above, likelihood-free parameter estimation has advanced, yet no effective framework exists for adaptively selecting informative actions in an online setting, as
has been done in likelihood-based approaches. To clarify the distinction between online
and offline estimation and to provide preliminaries for our proposed method, Section~\ref{sec:sec3}
presents the background of NPE with offline observations and explains why relying on offline data collection poses limitations in robotics.

\section{Preliminaries}\label{sec:sec3}
In this section, we first define the notation used for tuning a black-box simulator with offline data, and then briefly explain NPE. Finally, we discuss the limitations of offline data collection for tuning black-box simulators in robotics.

\subsection{Parameter tuning of black-box simulators with offline data}
Let $p(\bx | \btheta)$ denote a stochastic simulator where $\bx$ and $\btheta$ represent the observation and simulation parameters, respectively. Using $p(\bx | \btheta)$ as a black-box simulator means that we can control the properties of an environment by changing the value of $\btheta$ and obtain the output $\bx$ by running a simulation without knowing the density function. Parameter tuning is the inverse process of simulation, where, for a given $\bx^{\text{obs}}$, it tries to determine simulation parameters that reproduce $\bx^{\text{obs}}$ with the following Bayes' rule:
\begin{equation}
p(\btheta | \bx^{\text{obs}}) = \frac{p(\bx^{\text{obs}} | \btheta)p(\btheta)}{p(\bx^{\text{obs}})} = \frac{p(\bx^{\text{obs}} | \btheta)p(\btheta)}{\int p(\bx^{\text{obs}}|\btheta)p(\btheta) d\btheta}
\end{equation}
The likelihood $p(\bx | \btheta)$ is needed to calculate the posterior $p(\btheta | \bx^{\text{obs}})$. However, the likelihood is inaccessible in black-box simulators, making analytical calculation of the posterior challenging. 

\subsection{Neural posterior estimation (NPE)}

NPE provides likelihood-free solutions to this problem by using a neural network to train a posterior estimator instead of analytical calculations with exact likelihoods. A posterior estimator is a parametric model $q_\phi$ that returns a conditional density function $q_\phi(\btheta | \bx)$ over a pair of $(\btheta, \bx)$. While any distribution family can be used as a density function, we use a $k$-Mixture of Gaussians ($k$-MoGs),  following previous work \cite{Ramos2019-oy}, that is, 
\begin{equation}
q_\phi(\btheta |\bx) = \sum_{k=1}^{K} \pi_k \mathcal{N}(\btheta | \boldsymbol{\mu}_k, \boldsymbol{\Sigma}_k)
\end{equation}
where  $\pi_k, \boldsymbol{\mu}_k, \boldsymbol{\Sigma}$ denote the weight, mean, and covariance of the $k$-th Gaussian component, respectively. This model is trained using a neural network $f_\phi (\bx)$, which takes $\bx$ as input and outputs the elements of the $k$-MoGs. 
\begin{equation}
    \{ \pi_k, \boldsymbol{\mu}_k, \boldsymbol{\Sigma}_k \}_{k=1}^{K} = f_\phi (\bx) 
\end{equation}
By minimizing the loss function 
\begin{equation}
    \mathcal{L}(\phi) = - \sum_n \log q_\phi (\btheta_n | \bx_n)
\end{equation}
over a dataset $\{ (\btheta_n, \bx_n )\}_n$ sampled from $p(\btheta, \bx)$, the estimator $q_\phi (\btheta | \bx)$ converges to the true posterior $p(\btheta | \bx)$~\cite{Papamakarios2016-ki}. Note that  $\{ (\btheta_n, \bx_n )\}_n$ can be sampled using forward simulation, where we first sample   $\btheta_n \sim p(\btheta)$, and then $\bx_n \sim p(\bx|\btheta_n)$ by running a forward simulation.

\begin{figure*}[tb]
    \centering
    \begin{center}
    \includegraphics[width=1.0\hsize]{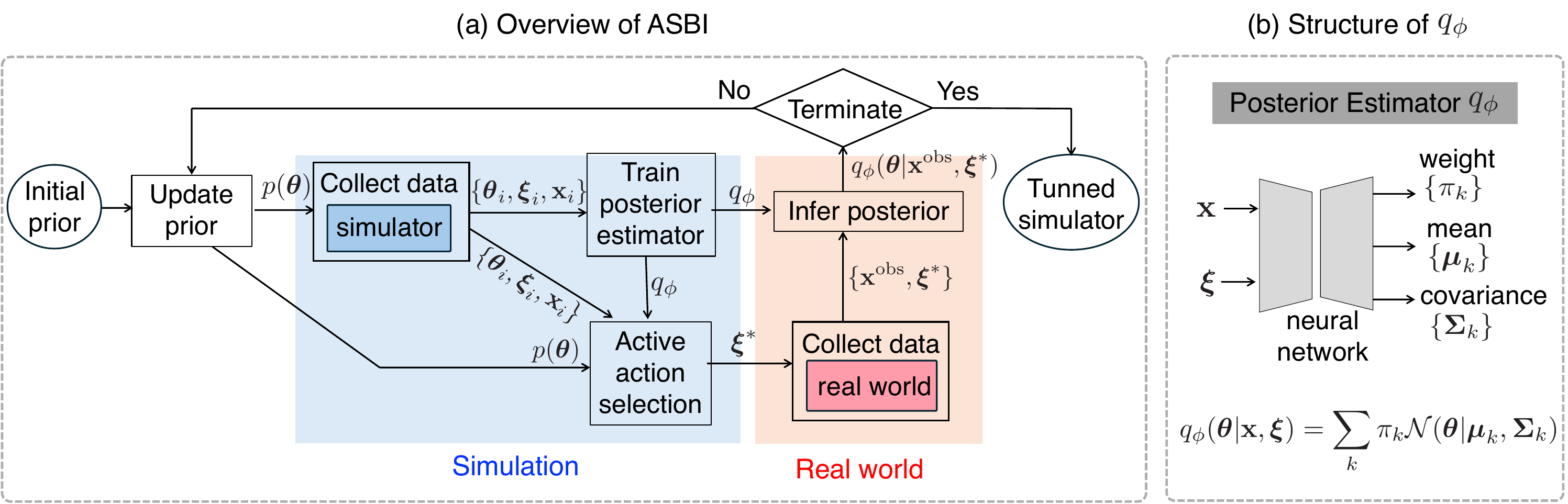}
    \end{center}
    \caption {(a) Overview of ASBI. ASBI enables accurate parameter estimation for black-box simulators by selecting the optimal action using forward simulation and a learned posterior estimator. (b) Structure of posterior estimator. To approximate the posterior without the likelihood, ASBI trains a posterior estimator using a neural network that takes observation and action as input and outputs a set of parameters for $k$-MoGs.}
    \label{fig:fig2}
\end{figure*}

\subsection{Limitations of parameter estimation with given offline data}

Robotics simulators often involve various parameters, and whether $\bx^{\text{obs}}$ contains useful information about the parameters varies depending on the types and range of parameters.
In black-box simulators, however, the relationships among parameters, actions, and observations are unknown, making it challenging to prepare effective offline data in advance. One potential solution involves collecting exhaustive offline data across all possible actions. However, this approach is impractical in robotics due to the high cost of real executions. 

For practical applications, we need to select a subset of actions that return informative observations. Since the effectiveness of an action varies depending on the prior knowledge of parameters, it is necessary to select actions adaptively instead of using a fixed offline selection. To address this issue, we propose an active parameter tuning framework that strategically selects the optimal action during the parameter tuning process.

\section{Proposed Method}\label{sec:sec4}

In our proposed method, we aim to estimate simulation parameters that replicate the real-world environment. In this study, we assume that the simulator and the real world have the same dynamics model, with no noise or model mismatch.

\subsection{Overview of active simulation-based inference (ASBI)}

We consider a robot action as a variable and optimize it in each round to collect informative online data from the real world. Let $\bxi$ denote a robot action. Both simulations and real-world observations are conditioned on $\bxi$:
\begin{equation}
    \bx \sim p(\bx | \btheta, \bxi), \ \ \ \ \bx^{\text{obs}} \sim \bar{p} (\bx | \bxi)
\end{equation}
where $p(\bx | \btheta, \bxi)$ represents the simulator, and $\bar{p}(\bx | \bxi)$ is the real-world environment we aim to reproduce in simulation. 

\vspace{10pt} 
\noindent{}{\bf Aim of the proposed method} In an online data collection setting, where the robot can collect real-world observations $\bx^{\text{obs}}$, the goal of ASBI is to determine the action $\bxi$ that leads to the most effective parameter estimation. We assume a myopic optimization scenario, 
meaning that the robot optimizes only one step ahead without planning over a long horizon. In this setting, the robot sequentially selects actions to iteratively collect additional data. 

Since the optimality of an action depends on the previous observation history, for example, repeatedly performing the same action does not provide new information, we adopt a sequential setting. In the following, we describe the sequential estimation process of ASBI. Detailed components are explained in Section~\ref{sec:sec4_2} and~\ref{sec:sec4_3}, and the entire procedure is summarized in Algorithm~\ref{alg:asbi}.

\vspace{10pt} 
\noindent{}{\bf Sequential process of ASBI} Let $\hat{p}(\btheta | \mathbb{D}_{r})$ denote the estimated posterior after round $r$, where $\mathbb{D}_{r}$ is the history of robot actions and observations from the real world. 
\begin{equation}
    \mathbb{D}_{r} = \{ (\bxi_1, \bx^{\text{obs}}_1), (\bxi_2, \bx^{\text{obs}}_2),
    \cdots, (\bxi_{r}, \bx^{\text{obs}}_{r})\} 
\end{equation}
where $\bxi_r$ is the robot action and $\bx^{\text{obs}}_{r}$ is
the corresponding observation obtained at round $r$, that is, $\bx^{\text{obs}}_{r} \sim \bar{p}(\bx | \bxi_{r})$. The sequential process for round $r+1$ proceeds as follows:

\begin{description}
  \item[{\bf (Step 1)}] {\bf Train posterior estimator with simulated data:} Collect a training dataset by running simulations with parameters sampled from the current prior, $\btheta \sim p(\btheta)$. Using this train dataset, retrain the posterior estimator $q_\phi$, a neural network that approximates the posterior distribution (see Section~\ref{sec:sec4_2} for details).

  \item[{\bf (Step 2)}] {\bf Action selection:} Determine the next optimal action \(\bxi^{*}_{r+1}\) by maximizing the objective function \(U(\bxi)\) defined in Section~\ref{sec:sec4_3}.  
  This objective approximates the information gain in a likelihood-free manner using the posterior estimator \(q_{\phi}\) and forward simulations.  

  \item[{\bf (Step 3)}] {\bf Data acquisition and posterior update:}
  Execute \(\bxi^{*}_{r+1}\) in the real world to obtain a new observation \(\bx^{\text{obs}}_{r+1}\).  
  Update the posterior by incorporating this new data:
  \[
    \hat{p}(\btheta \mid \mathbb{D}_{r+1}) = q_{\phi}\!\left(\btheta \mid \mathbf{x}^{\text{obs}}_{r+1}, \bxi^{*}_{r+1}\right).
  \]
  If the robot continues to collect data, set this posterior as the new prior for the next round. 
  \[ p(\btheta) \gets \hat{p}(\btheta \mid \mathbb{D}_{r+1}).
 \]
\end{description}

In Section~\ref{sec:sec4_2}, we explain how to extend NPE to handle action variables. Then, Section~\ref{sec:sec4_3} describes the objective function $U(\bxi)$, which determines the myopic action selection.

\begin{algorithm}[tb]
\caption{Active Simulation-Based Inference (ASBI)}\label{alg:asbi}
\begin{algorithmic}[1] 
    \Require Black-box simulator $p(\bx 
            | \btheta, \bxi)$, Target real-world environment $\bar{p}(\bx | \bxi)$, Discrete Action space $\mathcal{D}$ with $\bxi \in \mathcal{D}$, Initial prior $p(\btheta)$, Posterior estimator $q_\phi(\btheta | \bx, \bxi)$ (a neural network that approximates the posterior), Number of simulations per round $N$, number of rounds $R$, number of samples $M$ used for the finite-sum approximation of $U(\bxi)$
    \Ensure 
        Estimated posterior $\hat{p}(\btheta | \mathbb{D}_R)$ after $R$ rounds. 
    \State Initial dataset $\mathbb{D}_0 = \emptyset$
    \State Initialize prior $p(\btheta)$
    \For{$r = 1, \dots, R$} \Comment{Sequential round $r$}
        \Statex \textbf{(Step 1) Train posterior estimator with simulated data}
        \For{$n = 1, \dots, N$} \Comment{Generate training data under current prior}
            \State Sample an action $\bxi_n$ uniformly from $\mathcal{D}$
            \State Sample parameters $\btheta_n \sim p(\btheta)$
            \State Simulate observation $\bx_n \sim p(\bx | \btheta_n, \bxi_n)$
        \EndFor
        \State Train $q_\phi$ by minimizing the negative log-likelihood: \Comment{Eq.~\ref{eq:mdnn_loss} (Section~\ref{sec:sec4_2})}
         $$ \phi  \gets \argmin_\phi - \sum_{n=1}^{N} \log q_{\phi} (\btheta_n | \bx_n, \bxi_n)$$ 

        \Statex \textbf{(Step 2) Action selection using myopic objective} \Comment{Eq.~\ref{eq:finite_sum_utility} (Section~\ref{sec:sec4_3})}
        \State Approximate $U(\bxi)$ via a finite sum with $M$ samples, then select: 
         \[
        \bxi_r^* = \arg\max_{\bxi \in \mathcal{D}} U(\bxi)
        \]

        \Statex \textbf{(Step 3) Collect real data and update posterior}
        \State Execute $\bxi_r^*$ in the real world to obtain observation $\bx_r^{\text{obs}} \sim \bar{p}(\bx \mid \bxi_r^*)$
        \State Update dataset $\mathbb{D}_r \gets \mathbb{D}_{r-1} \cup \{(\bxi_r^*, \bx_r^{\text{obs}})\}$
        \State Update the posterior by inserting observation and action into $q_\phi$:
        \[
            \hat{p}(\btheta \mid \mathbb{D}_r) = q_\phi(\btheta \mid \bx_r^{\text{obs}}, \bxi_r^*)
        \]
        
        \If{$r \neq R$} 
            \State Update prior for the next round: 
                $p(\btheta) \gets \hat{p}(\btheta \mid \mathbb{D}_r)$
        \EndIf
    \EndFor 
\end{algorithmic}
\end{algorithm}

\subsection{Action-extended neural posterior estimator}\label{sec:sec4_2}

To estimate the posterior conditioned on both $\bx$ and $\bxi$, we consider the following neural network $f_\phi$, which takes a pair $(\bx, \bxi)$ as input and outputs the elements of $q_\phi(\btheta | \bx, \bxi)$. 
\begin{align}
    \label{eq:nn_output}
    f_\phi (\bx, \bxi) &=\{ \pi_k, \boldsymbol{\mu}_k, \boldsymbol{\Sigma}_k \}_{k=1}^{K}  \\
    \label{eq:post_estimator}
    q_\phi(\btheta |\bx, \bxi) &= \sum_{k=1}^{K} \pi_k \mathcal{N}(\btheta | \boldsymbol{\mu}_k, \boldsymbol{\Sigma}_k)
\end{align}
Then, $q_\phi(\btheta |\bx, \bxi)$ can approximate $p(\btheta | \bx, \bxi)$ if $f_\phi$ is trained to minimize the loss function
\begin{equation}\label{eq:mdnn_loss}
    \mathcal{L}(\phi) = - \sum_{n=1}^{N} \log q_\phi (\btheta_n | \bx_n, \bxi_n)
\end{equation}
with a training dataset sampled from $p(\btheta, \bx, \bxi)$. Samples from $p(\btheta, \bx, \bxi)$ can be obtained using forward simulation $p(\bx | \btheta, \bxi)$:
\begin{equation}
    \btheta \sim p(\btheta),\  \bxi \sim \text{Uniform}, \  \bx \sim p(\bx | \btheta, \bxi)
\end{equation}
where we assume $p(\btheta|\bxi)=p(\btheta)$. We refer to this estimator as an action-extended NPE.

\subsection{Utility function}\label{sec:sec4_3}

To evaluate the effectiveness of an action $\bxi$, we define a utility function $U(\bxi)$ motivated by the concept of information gain~\cite{Lindley1956-ge}. For a given action $\bxi$, the information gain is given by
\begin{equation}\label{eq:mi_origin}
  I(\btheta, \bx | \bxi) = \iint p(\btheta, \bx | \bxi) \log \frac{p(\btheta | \bx, \bxi)}{p(\btheta)} d\btheta d\bx
\end{equation}
where we assume $\btheta$ to be independent of $\bxi$. Intuitively, this measures the expected reduction in uncertainty between the prior $p(\btheta)$ and the true posterior $p(\btheta | \bx, \bxi)$. However, in likelihood-free settings, directly computing Eq.~\ref{eq:mi_origin} is challenging because the likelihood is unavailable. To address this, ASBI leverages the action-extended NPE, $q_\phi (\btheta | \bx, \bxi)$, which serves as an approximate posterior of the true posterior.

Accordingly, we define the utility function, which serves as the objective to be maximized, as follows:
\begin{align}
    U(\bxi) &:= \iint p(\btheta, \bx | \bxi) \log \frac{q_\phi(\btheta | \bx, \bxi)}{p(\btheta)} d\btheta d \bx
\end{align}
Since this integral is generally intractable, we approximate it using Monte Carlo sampling with forward simulations. Let $M$ denote the total number of samples drawn from $p(\btheta, \bx | \bxi)$. Since $p(\btheta | \bxi) = p(\btheta)$, the joint distribution factorizes as $p(\btheta, \bx | \bxi) = p(\bx | \btheta, \bxi)p(\btheta)$. Then, $U(\bxi)$ can be approximated via the following procedure:
\begin{itemize}
\item [(1)] Sample parameters $\btheta_i \sim p(\btheta)$ from the prior for $(i = 1, \cdots, M)$.
\item [(2)] Simulate observations $\bx_i \sim p(\bx | \btheta_i, \bxi)$ for the given action $\bxi$.
\item [(3)] Evaluate the posterior $q_\phi (\btheta_i | \bx_i, \bxi)$ by inserting the simulated data into the action-extended NPE.
\item[(4)] Approximate the expected information gain $U(\bxi)$ using a finite sum:
\begin{align}\label{eq:finite_sum_utility}
U(\bxi) \approx \frac{1}{M} \sum_{i=1}^M \log \frac{q_\phi(\btheta_i | \bx_i, \bxi)}{p(\btheta_i)}
\end{align}
\item [(5)] The optimal action is selected by maximizing $U(\bxi)$. 
\begin{equation}\label{eq:utility}
\bxi^{*} = \underset{\bxi} {\operatorname{argmax}}~U(\bxi)
\end{equation}
\end{itemize}

This likelihood-free formulation allows ASBI to evaluate and compare the expected information gain of different actions without requiring explicit likelihood models, making it suitable for black-box simulators where only forward sampling is available.

\section{Simulation Experiment}\label{sec:sec5}

In this section, we perform three sim-to-sim experiments to demonstrate the performance of ASBI.  As opposed to real-world experiments, sim-to-sim experiments estimate the parameters of virtual environments prepared using simulators with fixed parameter values. It is assumed that the true parameters are unobservable when estimating parameters and are used only for quantitative performance evaluation. First, we explain the baseline methods in Section~\ref{sec:sec5_1}. Then, we compare the performance of ASBI using a numerical model (Section~\ref{sec:sec5_2}) and a more complex robotics simulator for the box-collision task (Section~\ref{sec:sec5_3}) and the particle-parameter estimation task (Section~\ref{sec:sec5_4}).

\subsection{Comparison methods}\label{sec:sec5_1}
This section describes the baseline methods.
\begin{itemize}
    \item  {\textbf{Naive Simulation-Based Inference (NSBI)}: To evaluate the effectiveness of action optimization, we use NSBI, in which the action optimization is removed from ASBI. In each round, a randomly selected action is used to gather an observation.}
    \item {\textbf{Active Likelihood-Based Inference (ALHI)}:} To evaluate the effectiveness of likelihood-free approximation, this method uses surrogate likelihood models to evaluate information gain. A surrogate likelihood model is trained using Neural Likelihood Estimation (NLE)~\cite{Papamakarios2019-xp}, which leverages neural networks and simulation data, similar to NPE, but takes a pair of parameters and an action as input and outputs the likelihood of an observation. Using a learned likelihood estimator $q_\phi$, we can calculate the information gain based on the following relation:
    
\begin{equation}
    \log \frac{p(\btheta | \bx, \bxi)}{p(\btheta | \bxi)} = \log \frac{p(\bx | \btheta, \bxi)}{p(\bx | \bxi)} \approx \log \frac{q_{\phi}(\bx | \btheta, \bxi)}{p(\bx|\bxi)}
\end{equation}
Here,  $p(\bx | \bxi)$ can be approximated by 
\begin{equation}\label{eq:mi_nlh}
    p(\bx|\bxi) = \int p(\bx | \btheta, \bxi) p(\btheta) d\btheta \approx \frac{1}{N} \sum_{\btheta \sim  p(\btheta)} q_\phi(\bx | \btheta, \bxi).
\end{equation}

    \item {\textbf{Naive Likelihood-Based Inference (NLHI)}:} To evaluate the effectiveness of action optimization, we use NLHI, in which the action optimization is removed from ALHI. In each round, a randomly selected action is used to gather an observation. 

    \item {\textbf{Sequential BED (SeqBED)} \cite{Kleinegesse2021-zj}:} We find SeqBED to be the closest parameter tuning technique to ours for black-box simulators. Due to its excessive simulation cost, we only compare this method in the numerical model.
\end{itemize}

\begin{figure*}
    \centering 
    \includegraphics[width=1.0\hsize]{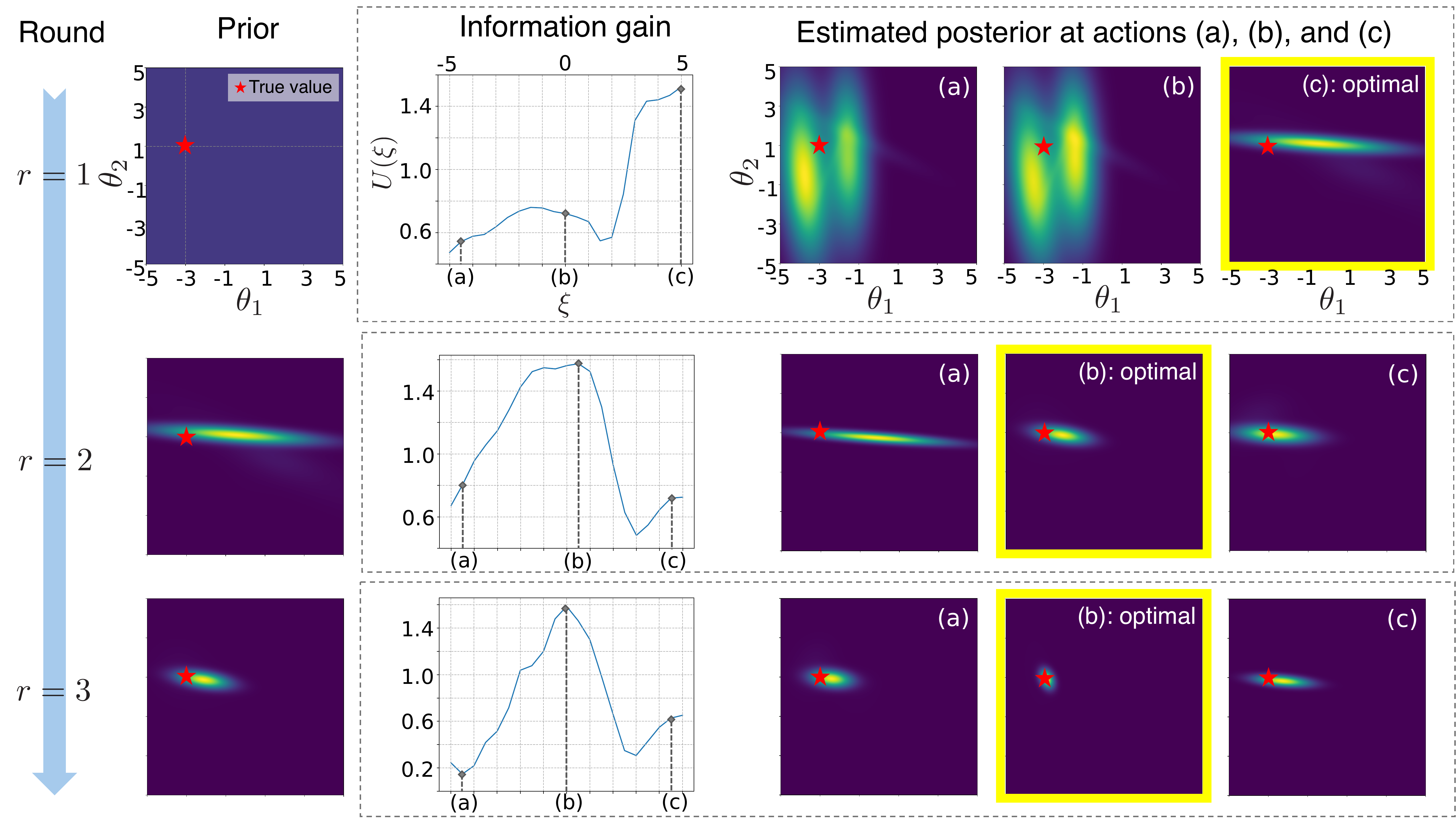}
    \caption{ Information gain graphs and comparison of estimated posterior at different actions obtained using ASBI for the numerical model. 
    Each row describes, for round $r$, the prior, the graph of information gain, and the posteriors obtained at different actions. Tested actions (a), (b), and (c) are shown with green diamonds on the information gain graph.  The posteriors on optimal actions  $(\xi^{*}_1, \xi^{*}_2, \xi^{*}_3) = (5.0, 0.5, 0.0)$ are highlighted in yellow, and each highlighted posterior is used as the prior of the next round.  The first round with action(c), $\xi^{*}_1=5.0$,  effectively reduces uncertainty in $\theta_2$ prediction. Then, in rounds 2 and 3, actions near zero reduce uncertainty in $\theta_1$ prediction. Using the informative observation obtained with the optimal action at each round, ASBI successfully estimates the posterior highly peaked around the true parameter $\btheta^{\text{true}} = (-3, 1)$, which is represented with a red star. }
    \label{fig:fig3}
\end{figure*}

\begin{figure}[tb]
\centering
\includegraphics[width=0.8\hsize]{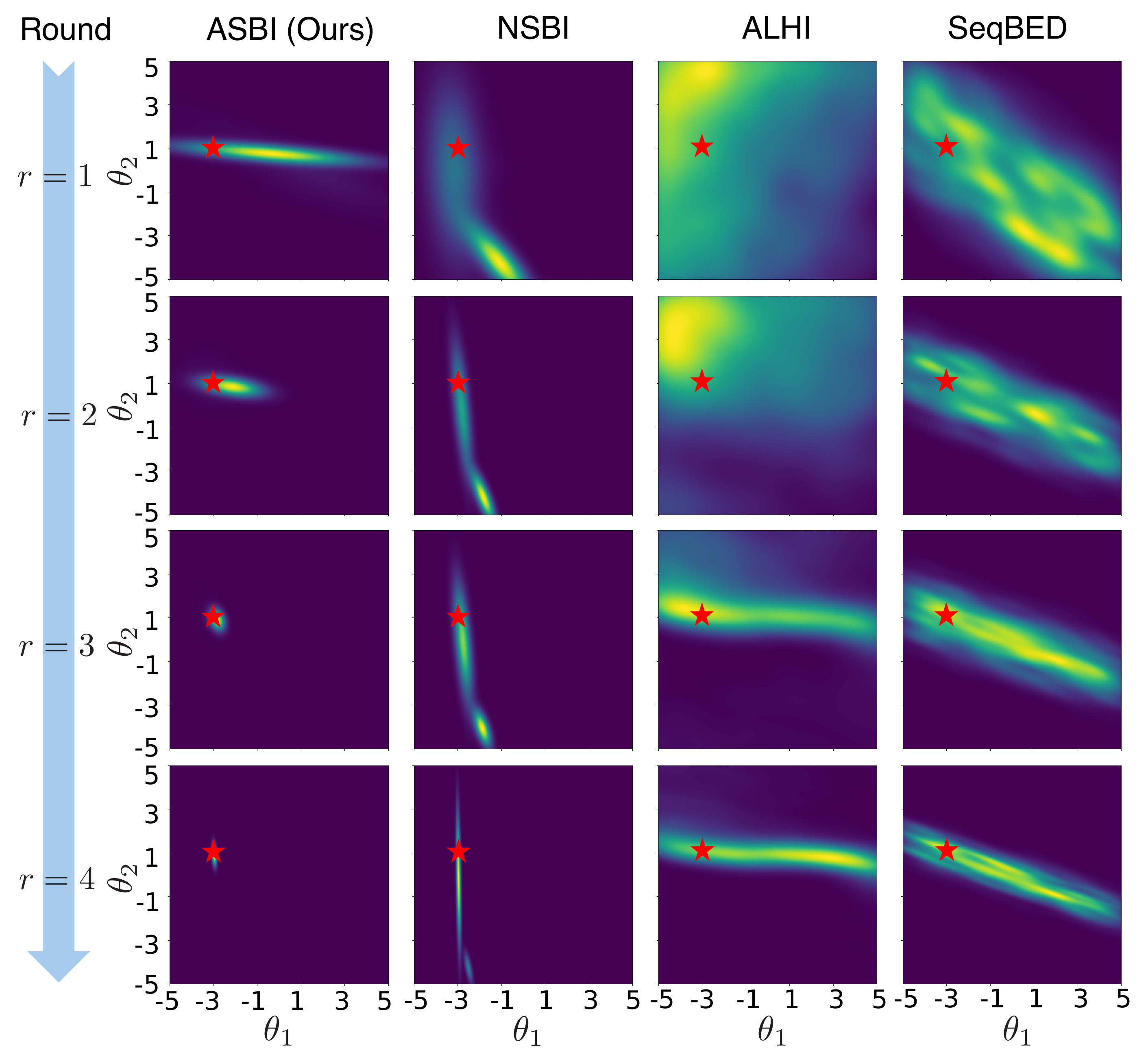}
\caption{ Comparison of the changes in posteriors through 4-round parameter estimation using different methods for the numerical model. The true parameters are shown with a red star.}
\label{fig:fig4}
\end{figure} 

\subsection{Sim-to-sim experiment with a numerical model}\label{sec:sec5_2} 

\noindent{}{\bf Simulation Model:}  We consider the following model with a two-dimensional parameter $\btheta = (\theta_1, \theta_2)$ and a one-dimensional action variable $\xi$. 
\begin{equation}\label{model:exp1}
    x = f_{\xi} (\btheta) = \theta_1 \exp (3-\xi) + \theta_2 \xi + \epsilon
\end{equation}
where a noise is drawn from the normal distribution $N(\epsilon; 0, 1)$. For given $\xi$ and $x$, the upper bound of parameter estimation is given as a linear relationship between $\theta_1$ and $\theta_2$. 

Note that we use this model as a black-box simulator that takes $(\theta_1, \theta_2, \xi)$ as input and outputs $x$.

\begin{figure}[tb]
\centering
\includegraphics[width=0.5\hsize]{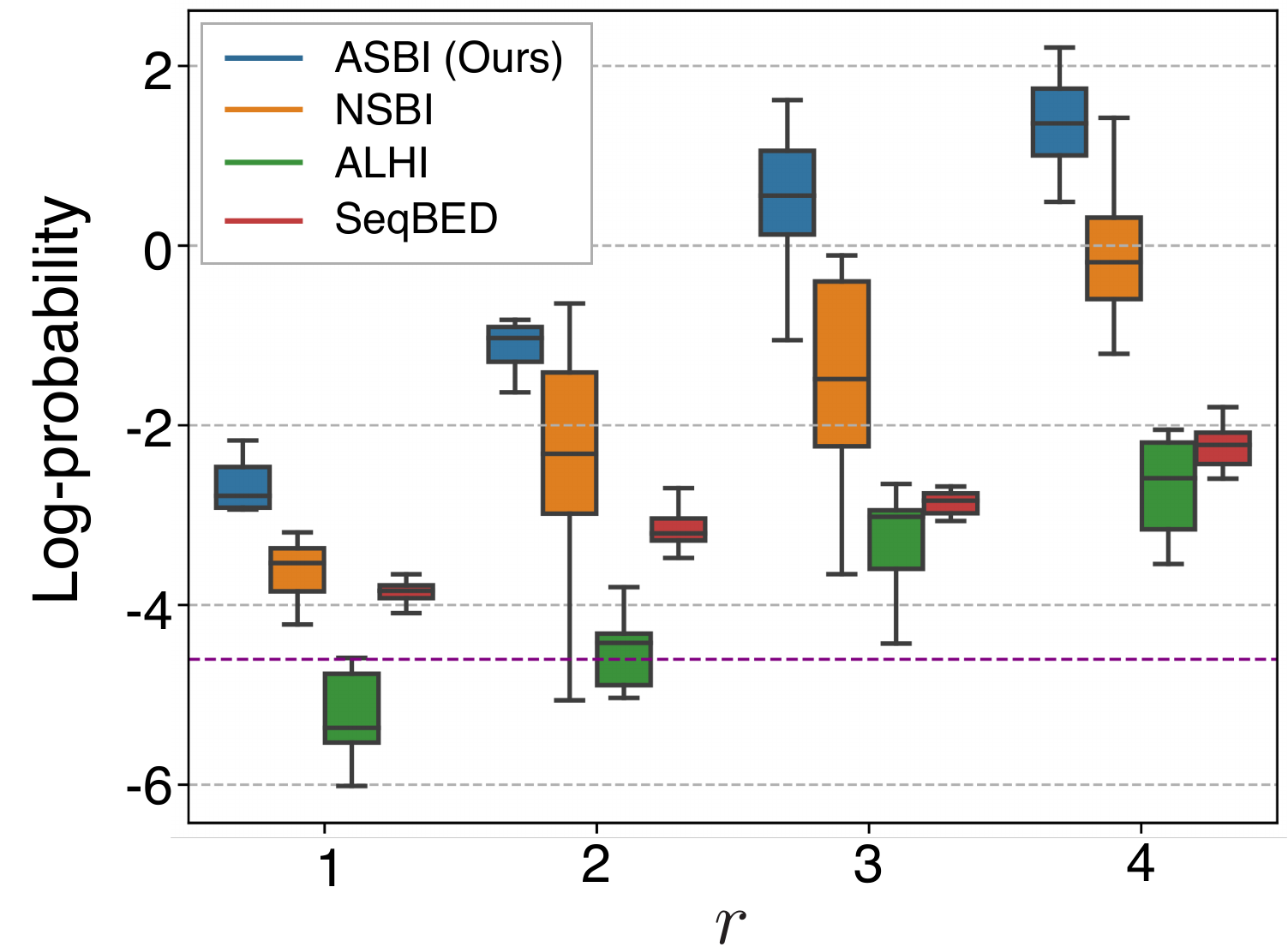}
\caption{ Comparison of log-probability of ground-truth parameter at each round $r$ for the numerical model. Log-probability of the initial prior $p(\btheta) = 0.01$ is shown with a dotted purple line.}
\label{fig:fig5}
\end{figure}

\noindent{}{\bf Simulation Parameter:} We assume that both $\theta_1$ and $\theta_2$ take continuous values in the interval $[-5, 5] $. The initial prior is assumed as a uniform distribution over this area. As a target environment to estimate, we use the simulator with $\btheta^{\text{true}}=(-3, 1)$. Assuming that the true parameter $\btheta^{\text{true}}$ is not observable, the goal is to estimate the parameter from observations collected interactively from the target environment.

\noindent{}{\bf Action Variable:} We assume $\xi$ is from the discrete space 
\begin{equation*}
\mathcal{D}=\{ -5.0, -4.5, -4.0, \cdots, 4.0, 4.5, 5.0\}.
\end{equation*}

\noindent{}{\bf Neural Posterior:} We use a neural network with two hidden layers of size (128, 128), which takes a two-dimensional vector $(x, \xi)$ as input and returns elements of $5$-MoGs. In each round, the neural network is trained in 2000 iterations with a batch size of 50.

\noindent{}{\bf Utility Calculation:} We calculate the utility function $U(\xi)$ of Eq.~\ref{eq:finite_sum_utility} using $M=1000$ samples in each round.

\begin{table}[tb]
    \small
    \centering 
        \caption{Comparison of $RepErr(\xi)$ measuring the error between the true output and reproduced outputs from parameters sampled from the final posteriors.}\label{tab:exp1}
    \begin{tabular}{cccc}
    \toprule 
     & $\xi=-3.0$ & $\xi=0.0$  & $\xi=3.0$ \\
     \midrule 
      ASBI (Ours) & \bf{34.5 $\pm$  10.7} & \bf{1.95 $\pm$  0.49}  & \bf{1.60 $\pm$ 0.36} \\
      NSBI & 127 $\pm$ 76 & 6.35 $\pm$ 3.74 & 2.01 $\pm$ 1.56  \\
      ALHI  & 940 $\pm$ 556  & 46.7 $\pm$ 27.6 & 4.35 $\pm$ 1.32  \\ 
      SeqBED  & 895 $\pm$ 264  & 44.4 $\pm$ 13.2 & 2.92 $\pm$ 1.44 \\
     \bottomrule 
    \end{tabular}
\end{table}

\noindent{}{\bf Evaluation:} We perform a 4-round estimation to obtain the final posterior. For NSBI, we use a neural network with the same structure as ASBI. ALHI uses a neural network with the same hidden layer of size (128, 128) but takes a pair $(\theta_1, \theta_2, \xi)$ as input and outputs elements of $5$-MoGs. For ALHI and SeqBED, Kernel Density Estimation (KDE) is used to approximate the posterior. For each method, we run a 4-round parameter estimation ten times. We measure the log probability of the true parameter, $\log q_\phi(\btheta^{\text{true}}|\xi^{*}_r , x^{\text{obs}}_r)
$, in each round $r$. \par

For evaluation, we use two complementary indicators. First, we measure the log-probability of the true parameter
\begin{equation}
    \log q_\phi(\btheta^{\text{true}}|\xi^{*}_r , x^{\text{obs}}_r)
\end{equation}
at each round $r$. This metric evaluates how well the estimated posterior concentrates around the ground-truth parameter, serving as a direct indicator of probabilistic inference quality. A higher log-probability indicates that the model assigns greater confidence to the true parameter, which is a standard and widely used criterion in parameter estimation. \par 

Second, we measure the Reproduction Error ($RepErr$), which measures the discrepancy between the true observations and the outputs reproduced by parameters sampled from the estimated posterior. For action $\xi$, we define $RepErr(\xi)$  as 
\begin{equation}\label{eq:rep_err}
    RepErr(\xi) = \frac{1}{N}  \sum_{i=1}^{N} \mathrm{abs} ( f_{\xi} (\theta_i)  - x^{\text{true}}|_{\xi})
\end{equation}
where $x^{\text{true}}|_{\xi}$  is the simulation output at $\theta^{\text{true}}$ and $\xi$, with the noise $\epsilon=0$. We sample $N=10^5$ numbers of $\theta$ from each estimated posterior to compute this value. This metric quantifies the discrepancy between the calibrated simulator and the real world, with lower values indicating better calibration. \par

Together, these two indicators capture both probabilistic accuracy (log-probability) and simulation-level consistency ($RepErr$), providing a comprehensive evaluation of the inference quality.

\noindent{}{\bf Results:} 
Fig.~\ref{fig:fig3} shows the estimation results (prior, graph of information gain, and posterior) of the first three rounds using ASBI. The optimal action selected using ASBI is $(\xi^{*}_1,\xi^{*}_2, \xi^{*}_3, \xi^{*}_4)=(5.0, 0.5, 0.0, -0.5)$, where it first reduces the uncertainty in $\theta_1$ with $\xi=5.0$ as shown in the first row of Fig.~\ref{fig:fig3}. Then, it reduces the uncertainty in $\theta_2$ with actions near zero. We also show the posteriors obtained using non-optimal actions in Fig.~\ref{fig:fig3}, indicating that the posterior with the optimal action successfully narrows down the predictive distribution compared to non-optimal actions. 

Fig.~\ref{fig:fig4} shows the posteriors obtained at each round using different methods. The selected actions $(\xi^{*}_1,\xi^{*}_2, \xi^{*}_3, \xi^{*}_4)$ for comparison methods are $(1.5, 1.0, -4.5, -0.5)$ for NSBI, $(5.0, 4.5, 4.5, 4.5)$ for ALHI, and $(3.0, 2.5, 2.5, 2.5)$ for SeqBED. While all methods gradually narrow down the posteriors, ASBI effectively updates the posterior so that it peaks around the true parameters with a small number of observations. 

Fig.~\ref{fig:fig5} shows the log probabilities of the ground-truth parameter across rounds for all methods as boxplots. Before any updates ($r=0$), the initial prior is a uniform distribution over the area $[-5, 5] \times [-5, 5]$, plotted as the dotted purple line at $\log p(\theta^{\text{true}}) = -4.61$. After just one round $(r=1)$, ASBI already improves beyond this uninformed baseline, achieving a median log-probability of $-2.78$. As sequential updates progress, ASBI exhibits the fastest posterior convergence, reaching a median log-probability of $1.35$ in round 4, which is the highest among all methods. Furthermore, the interquartile range (IQR) for ASBI remains relatively tight throughout the rounds, indicating more consistent and stable inference, whereas NSBI and ALHI maintain a wider IQR, showing higher variability. 

Next, Table~\ref{tab:exp1}  shows how close the reproduced outputs are to the true output.
Reproduction Error ($RepErr$) was evaluated for three representative actions $\xi=(-3.0, 0.0, 3.0)$. ASBI consistently achieves the lowest reproduction error with minimal
variance across all actions. At $\xi=-3.0$, ASBI reduces the error by around 73\% compared
to NSBI, and by more than 90\% compared to SeqBED and ALHI. For $\xi = 0.0$, ASBI maintains
approximately 69\% lower error than NSBI and achieves over 90\% lower error relative to
SeqBED and ALHI. For $\xi=3.0$, ASBI reduces the error by about 20\% over NSBI, roughly 63\% over ALHI, and 45\% over SeqBED. Also, ASBI achieves the lowest standard deviation across all
actions, demonstrating that ASBI not only improves average calibration accuracy but also
significantly reduces variability, resulting in more accurate and stable parameter estimation compared to all other baseline methods.


\subsection{Sim-to-sim experiment with robotics simulator on box-collision task}\label{sec:sec5_3}

\begin{figure}[tb]
\centering
\includegraphics[width=0.75\hsize]{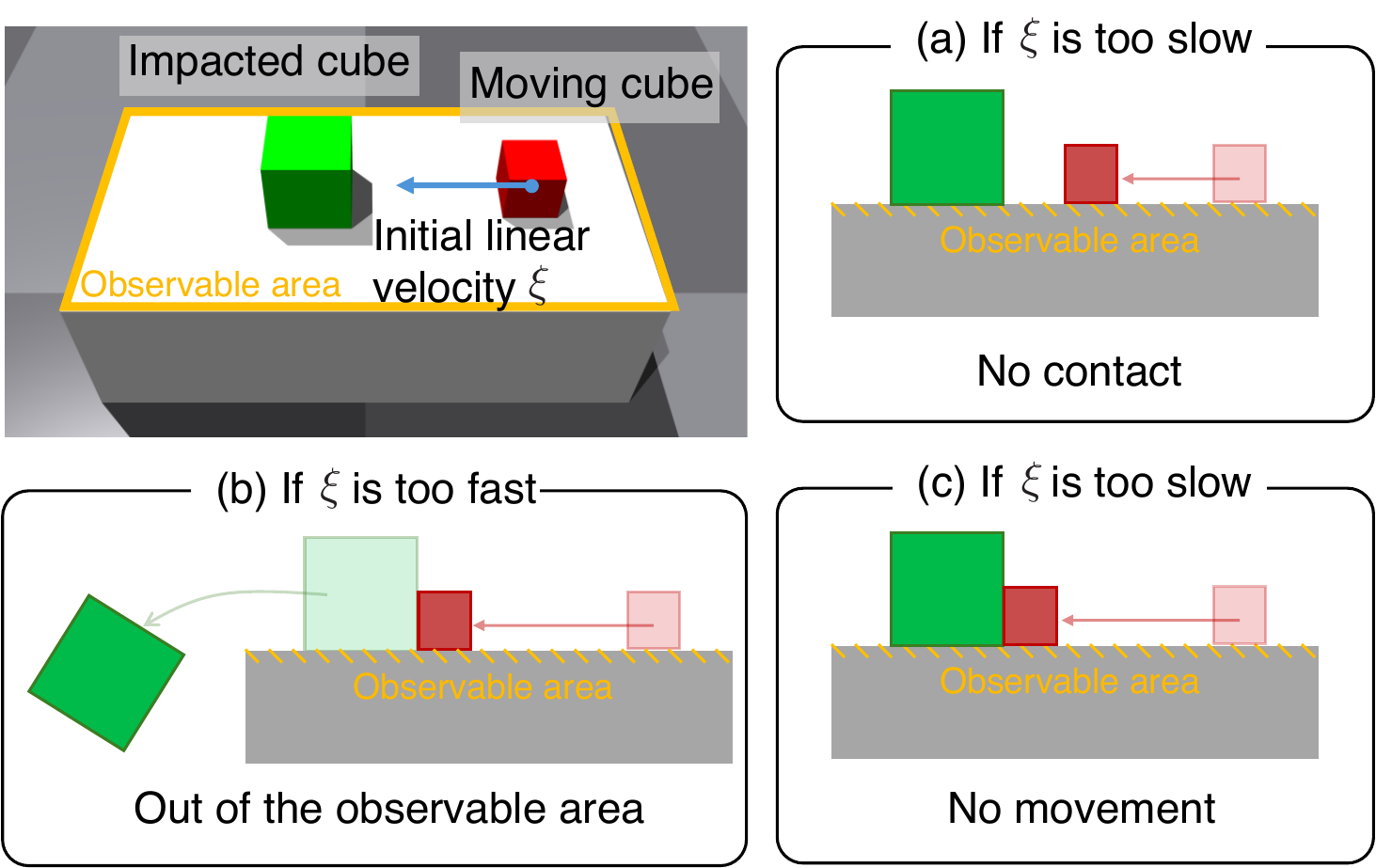}
\caption{Simulation setup for the Box-Collision Task. The moving cube (pusher, red) is assigned an initial linear velocity $\xi$ toward the stationary cube (impacted, green). Depending on the chosen velocity, the impacted cube may remain unmoved ((a), (c)) or leave the table boundary (b), highlighting the importance of proper action selection. }
\label{fig:fig6}
\end{figure}

Next, we evaluate ASBI in a more complex setting with nonlinear contact dynamics using a robot-oriented black-box simulator. In this experiment, we consider a box-collision scenario involving two cubic objects, as shown in Fig.~\ref{fig:fig6}. The goal is to estimate unknown physical parameters from the observed displacement of the impacted cube while optimizing the initial linear velocity of the moving cube.

\noindent{}{\bf Simulation Model:} In the initial frame, two cubic objects are placed on a flat table surface. The moving cube (pusher) has a size of 10 cm and a density of $1000$ kg/m$^{3}$. The stationary cube (impacted) has a size of 15 cm and an unknown density ranging from $0$ kg/m$^{3}$ to $1000$ kg/m$^{3}$. At each round, the moving cube is assigned an initial linear velocity $\xi$ m/s directed toward the stationary cube.  

After the collision, only the position of the impacted cube at a predefined timestep $t=100$ is recorded as the observation $\mathbf{x}$, where $\mathbf{x}=(x_0, x_1, x_2)$ represents its 3D position. If the impacted cube moves beyond the table boundary at the observation time, the measurement is considered invalid and encoded as $(-1,-1,-1)$. 

Compared to the numerical model in Section~\ref{sec:sec5_2}, this collision scenario involves nonlinear contact dynamics and further restricts the valid observation space, making the inference problem significantly more challenging. Moreover, this setup better reflects real robotic environments, where sequential online data is actively collected based on robot-object interactions. 

\noindent{}{\bf Simulation Parameter:} We estimate the three-dimensional parameter $\btheta=(\theta_{tf}, \theta_{cf}, \theta_{cd})$, where $\theta_{tf}$ is the coefficient of friction of the table, $\theta_{cf}$ the coefficient of friction of the impacted cube, and $\theta_{cd}$ the density of the impacted cube, which range from 0 to 1. While the actual density ranges from 0 to 1000, it is rescaled to $[0, 1]$, that is, $\theta_{cd}=0.5$ describes a density of $500$ kg/m$^{3}$. The initial prior is assumed as a uniform distribution over this area. 

\noindent{}{\bf Action Variables:} The action $\xi$ corresponds to the initial velocity of the moving cube. If $\xi$ is too small, the impacted cube does not move. If $\xi$ is too large, the impacted cube falls off the table and the observation becomes invalid $(-1,-1,-1)$. Thus, selecting an informative velocity is critical for reducing parameter uncertainty.  We assume $\xi$ is from the discrete space 
\begin{equation}
    \mathcal{D} = \{ 0.0, \ 0.5, \ 1.0, \ 1.5, \cdots , \ 18.5, \ 19.0, \ 19.5, \ 20.0 \}
\end{equation}
where the unit is m/s.

\begin{figure}[bt]
    \centering
    \includegraphics[width=0.90\columnwidth]{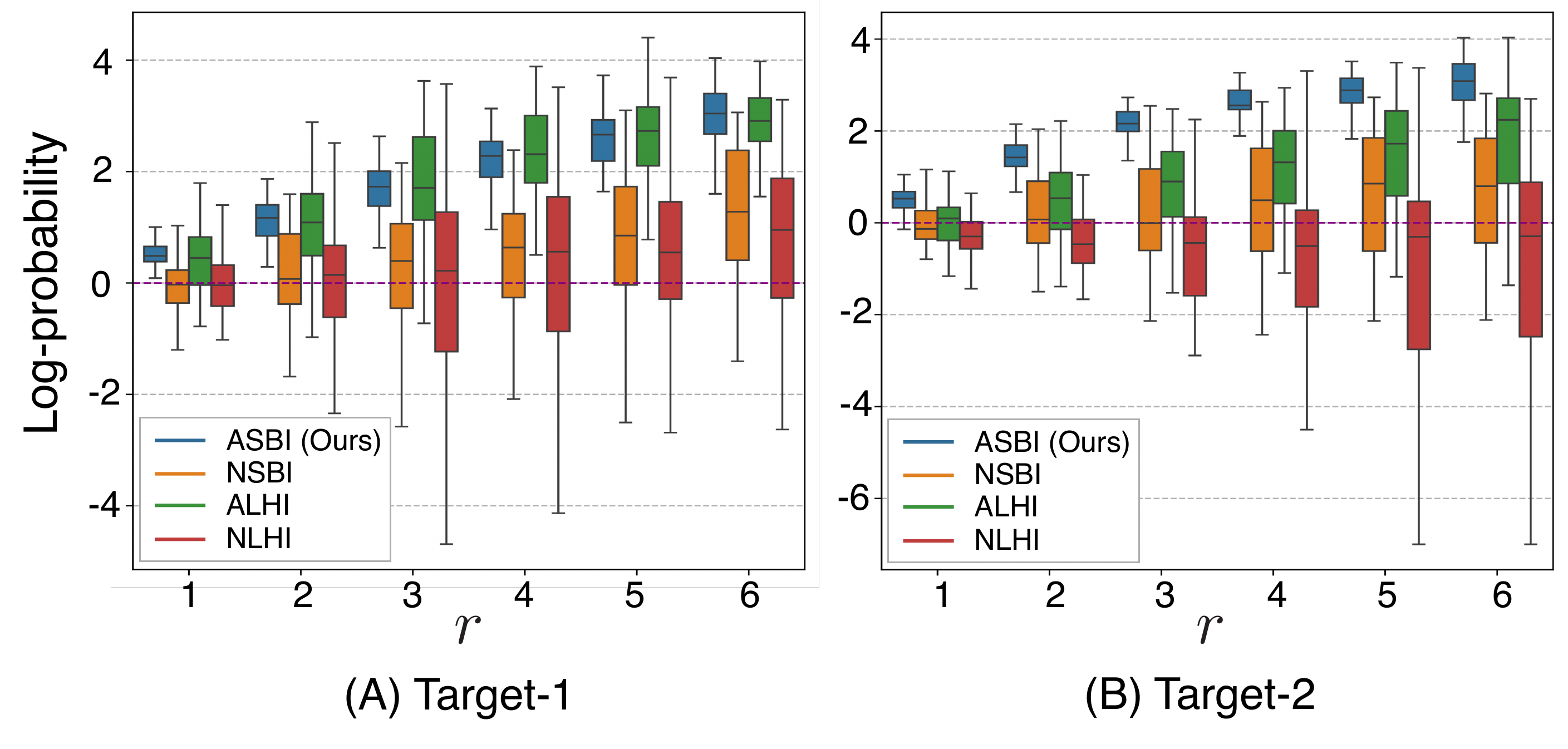}
    \caption{
        Log-probabilities of the ground-truth parameters across rounds for the box-collision task. ASBI consistently achieves the highest median and tightest IQR. Notably, ASBI achieves high accuracy from early rounds, demonstrating its ability to estimate parameters effectively with only a few observations. 
    }\label{fig:fig7}
\end{figure}




\begin{sidewaystable}
\centering
\caption{Reproduction Error ($RepErr$) of sim-to-sim experiment for box-collision. Mean $\pm$ std over 50 trials. \textbf{All values are scaled by $10^{-3}$ and rounded to one decimal place.}}
\label{tab:exp2}
\small
\setlength{\tabcolsep}{6pt}
\renewcommand{\arraystretch}{1.2}

\begin{tabular*}{\textwidth}{@{\extracolsep{\fill}}lcccccccc}
\toprule
& \multicolumn{4}{c}{\textbf{Target-1}} & \multicolumn{4}{c}{\textbf{Target-2}} \\
\cmidrule(lr){2-5} \cmidrule(lr){6-9}
\textbf{Action} & \textbf{ASBI (Ours)} & \textbf{NSBI} & \textbf{ALHI} & \textbf{NLHI} & \textbf{ASBI (Ours)} & \textbf{NSBI} & \textbf{ALHI} & \textbf{NLHI} \\
\midrule
$\xi = 3$ & \textbf{2.7 $\pm$ 0.6} & 19.5 $\pm$ 23.0 & 4.4 $\pm$ 4.3 & 42.5 $\pm$ 66.0 &
\textbf{703 $\pm$ 190} & 1850 $\pm$ 250 & 1350 $\pm$ 580 & 1930 $\pm$ 180 \\

$\xi = 5$ & \textbf{10.3 $\pm$ 2.5} & 158 $\pm$ 230 & 20.5 $\pm$ 20.0 & 297 $\pm$ 360 &
\textbf{0.0 $\pm$ 0.0} & 477 $\pm$ 540 & 262 $\pm$ 440 & 851 $\pm$ 570 \\

$\xi = 7$ & \textbf{22.7 $\pm$ 8.7} & 532 $\pm$ 590 & 84.5 $\pm$ 130 & 754 $\pm$ 650 &
\textbf{0.0 $\pm$ 0.0} & 77.0 $\pm$ 140 & 59.5 $\pm$ 160 & 300 $\pm$ 370 \\

\bottomrule
\end{tabular*}
\end{sidewaystable}

\noindent{}{\bf Network Training:} We use a neural network with two hidden layers of size (128, 128), which takes a four-dimensional input vector $(x_0, x_1, x_2, \xi)$ and outputs elements of 5-MoGs. This experiment performs a 6-round of sequential parameter estimation. In each round, 500 parameters are sampled from the current prior for each action, and simulations are run to generate the training dataset. Each round trains the network for 2000 iterations with a batch size of 50, and the resulting posterior becomes the prior for the next round.

\noindent{}{\bf Utility Calculation:} We calculate the utility Eq.~\ref{eq:finite_sum_utility} with $M=250$ samples from the training dataset of neural networks. 

\noindent{}{\bf Evaluation:} For quantitative evaluation, we use simulation as the target environment with two distinct values. Target-1 represents a high-friction, high-density scenario with the ground-truth parameters set to $\btheta^{\text{true}}=(\theta_{tf}^{\text{true}}, \theta_{cf}^{\text{true}}, \theta_{cd}^{\text{true}}) = (0.8, 0.8, 0.8)$, while Target-2 represents a low-friction, low density scenario with $\btheta^{\text{true}}=(\theta_{tf}^{\text{true}}, \theta_{cf}^{\text{true}}, \theta_{cd}^{\text{true}}) = (0.2, 0.2, 0.2)$. 

We compare ASBI against three baselines: NSBI (ASBI without action optimization, i.e., random action), ALHI (active likelihood-based inference), and NLHI (ALHI without action optimization, i.e., random actioin). For each method, we perform six sequential rounds of parameter estimation, repeating the process 50 times for each method. 

Performance is evaluated in two metrics. First, we measure the log-prbability of the ground-truth parameter, 
$$\log q_\phi (\boldsymbol{\theta}^{\text{true}} | \xi^{*}_r, x^{\text{obs}}_r)$$
at each round $r$ from 1 to 6. Second, we measure the reproduction accuracy, $RepErr$, defined in Eq.~\ref{eq:rep_err}, evaluated for representative actions $\xi=3, 5, 7$. Here, $RepErr$ quantifies the averaged Euclidean distance between the impacted cube's position in the calibrated simulation and the true observation, averaged over $N=250$ samples from the estimated posterior.

\noindent{}{\bf Results:} Fig.~\ref{fig:fig7} shows the log-probabilities of the ground-truth parameters across six rounds for two targets, Target-1 and Target-2. Starting from a uniform prior $\log p(\boldsymbol{\theta}^{\text{true}})=0.0$, ASBI improves beyond the uninformed baseline and continues to converge rapidly, achieving the highest median log-probability by the final round. Compared to NSBI, ALHI, and NLHI, ASBI achieves not only higher median values but also tighter interquartile ranges, indicating more stable and consistent inference.  

Table~\ref{tab:exp2} shows $RepErr$ for representative actions $\xi=3$, $5$, and $7$. Across all settings, ASBI consistently achieves the lowest $RepErr$ for both Target-1 and Target-2. For instance, at $\xi = 3$, ASBI yields more than 47\% lower error than ALHI and over 60\% lower error than NSBI for Target-2. At $\xi=5$ and $\xi=7$, ASBI achieves zero error, as the impacted cube falls off the table in both the calibrated simulations and the ground-truth observation, resulting in perfectly matched outcomes.

\noindent{}{\bf Ablation Analysis:} To further clarify the contribution of each component in ASBI, we conducted an ablation study separating the benefits of direct posterior estimation and adaptive action selection. 

First, the effect of action-extended NPE is examined by comparing ASBI with ALHI,  and NSBI with NLHI. As shown in Fig.~\ref{fig:fig7} and Table~\ref{tab:exp2}, posterior-based methods consistently achieve higher log-probabilities and lower $RepErr$ compared to their likelihood-based counterparts, demonstrating that directly learning the posterior avoids instability in surrogate likelihood estimation. 

Second, the role of adaptive action selection is evaluated by comparing ASBI with NSBI in posterior-based frameworks and ALHI with NLHI in likelihood-based frameworks. Action optimization significantly accelerates posterior convergence from early rounds, reduces invalid observations, and improves estimation accuracy. This confirms that our active action selection strategy benefits both posterior-based and likelihood-based approaches.

These ablation results show that both components contribute complementary benefits: direct posterior estimation improves inference accuracy and stability, while adaptive action selection enhances sample efficiency by focusing exploration on informative actions. Together, they enable ASBI to outperform all baselines in both accuracy and robustness.

\begin{figure}[tb]
    \centering
    \includegraphics[width=0.90\columnwidth]{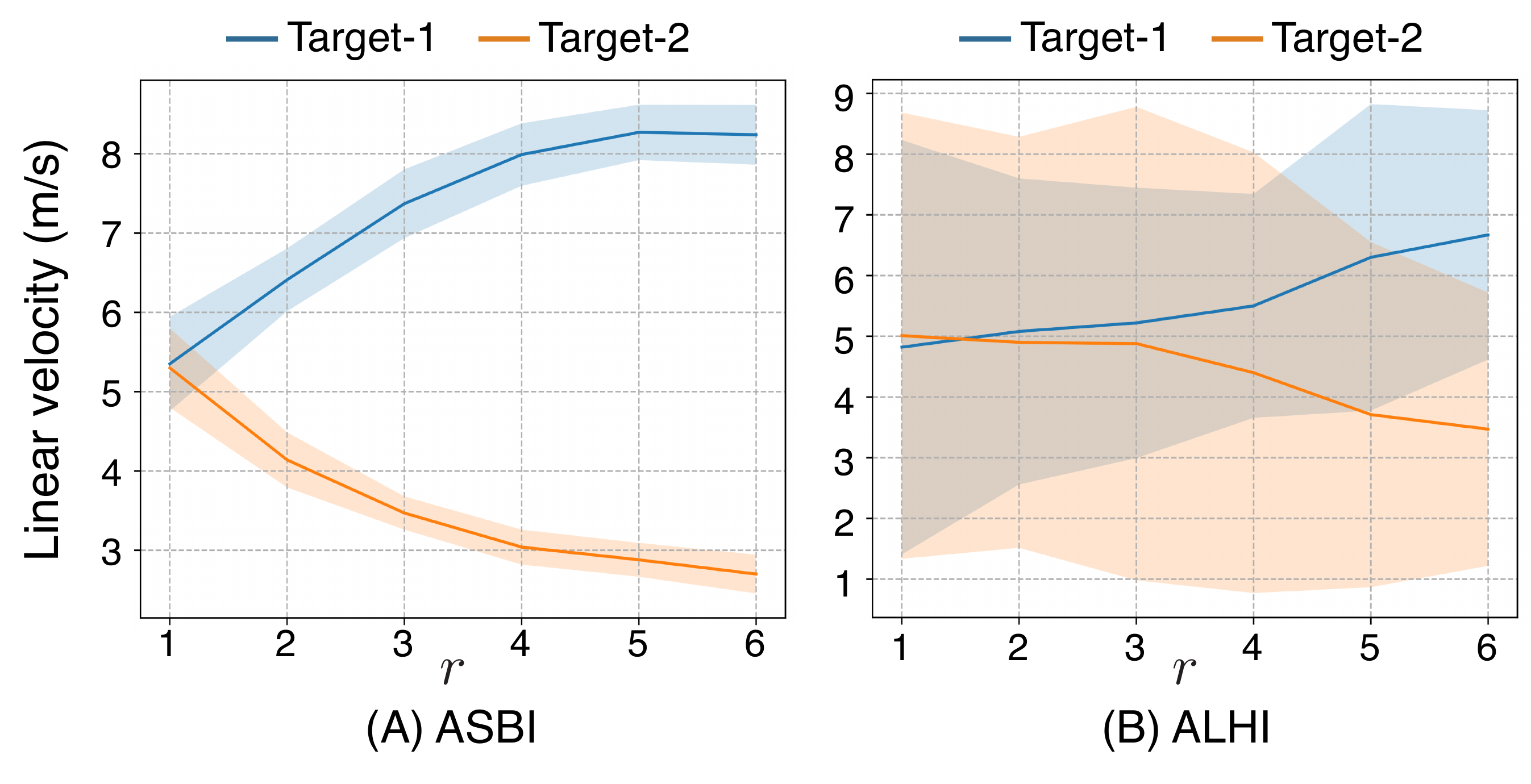}
    \caption{
        Selected actions across rounds in the box-collision task for ASBI (left) and ALHI (right). The blue line represents Target-1 (high friction and high density), while the orange line represents Target-2 (low friction and low density). ASBI shows clear adaptive trends with stable and consistent action choices, whereas ALHI exhibits inconsistent choices with high variance.
    }\label{fig:fig8}
\end{figure}

\noindent{}{\bf Action Selection Analysis:} 
Fig.~\ref{fig:fig8} visualizes the selected actions across rounds. ASBI exhibits a clear adaptive pattern with low variance. For Target-1 (high friction and high density), it gradually increases selected velocities from 5.35 m/s in round 1 to about 8.2 m/s in round 6, matching the intuitive need for stronger pushes. Conversely, for Target-2 (low-friction and low-density), it progressively decreases velocity from 5.3 m/s in round 1 to about 2.7 m/s in round 6, reflecting the need for gentler pushes to avoid invalid observations beyond the table boundary. In contrast, ALHI shows weak adaptive tendencies and high variance, indicating unstable exploration. These intuitive and stable patterns in ASBI correlate directly with its more stable posterior updates and lower $RepErr$, confirming that active action selection of ASBI is a key factor in improving inference stability and accuracy. 

\begin{figure}[H]
    \centering
    \includegraphics[width=0.50\columnwidth]{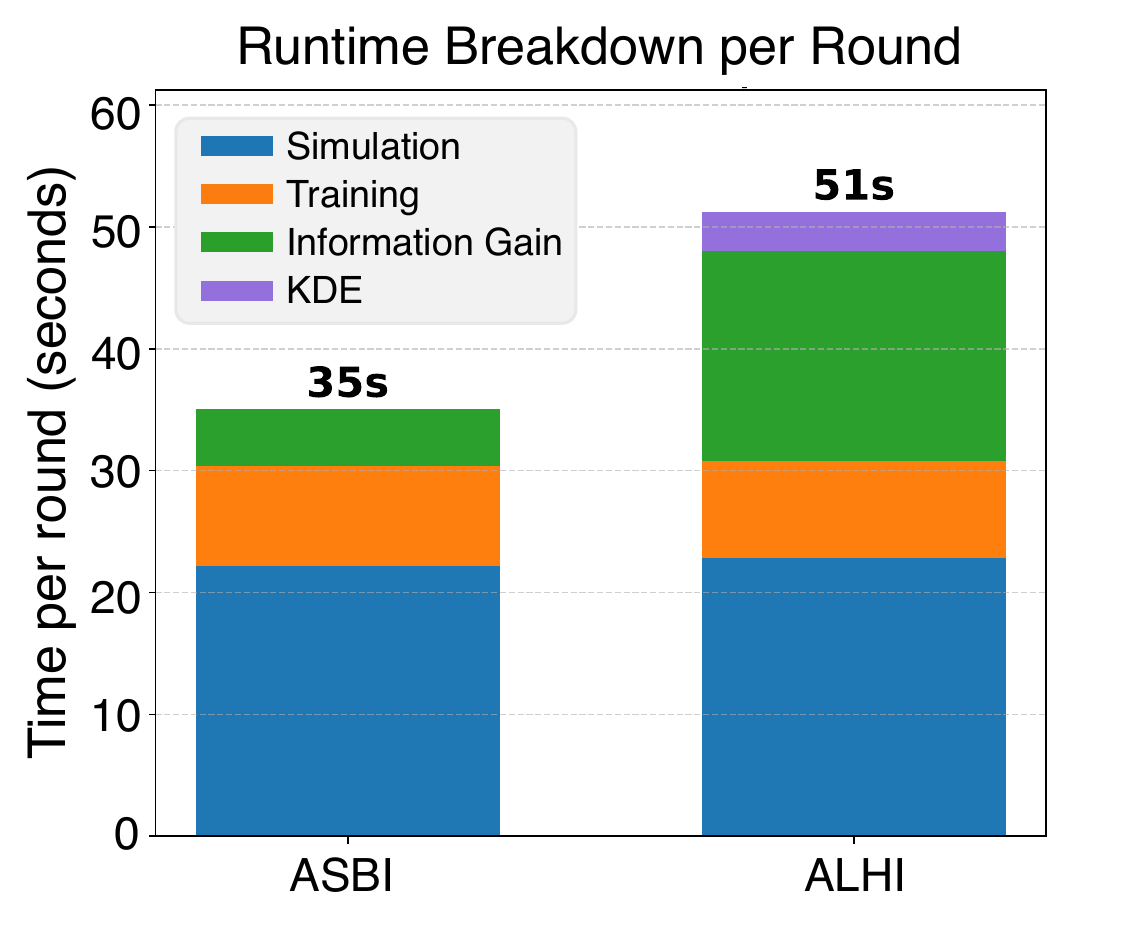}
    \caption{
       Average runtime breakdown per round for ASBI and ALHI in the box-collision task. Runtime is divided into four parts: (blue) simulation, (orange) training of the neural network, (green) calculation of information gain, and (purple) KDE fitting for obtaining the posterior. This is required only for ALHI.}
    \label{fig:fig9}
\end{figure}

\noindent{}{\bf Computation Time Analysis:} Fig.~\ref{fig:fig9} shows the runtime breakdown per round for ASBI and the likelihood-based baseline ALHI in the box-collision task, averaged over 100 trials of Target-1 and Target-2. All experiments were conducted on a workstation running Ubuntu 22.04 LTS, equipped with an Intel Core i9-11900K CPU (3.5 GHz), 64 GB RAM, and an NVIDIA GeForce RTX 3080 Ti GPU (12 GB VRAM). While simulations were executed on the GPU, all other processes were performed on the CPU. The implementation was developed in Python 3.8.16, using PyTorch 2.0.1+cu117.

The total runtime is decomposed into three components for ASBI: simulation, neural network training, and information gain calculation, and four components for ALHI, which requires an additional kernel density estimation (KDE) step to recover the posterior from the surrogate likelihood. 

On average, ASBI required 22.2 s for simulation, 8.2 s for training, and 4.6 s for information gain calculation, resulting in a total of approximately 35 s per round. In contrast, ALHI required 22.8 s for simulation and 7.9 s for training, but a much higher 17.2 s for information gain calculation plus an additional 3.2 s for KDE fitting, totaling roughly 51 seconds per round. This runtime comparison shows that ASBI reduces the cost of computing information gain by approximately 73\% compared to ALHI, as it directly computes the posterior without relying on surrogate likelihood estimation.


\begin{figure}[tb]
\centering 
\includegraphics[width=0.8\columnwidth]{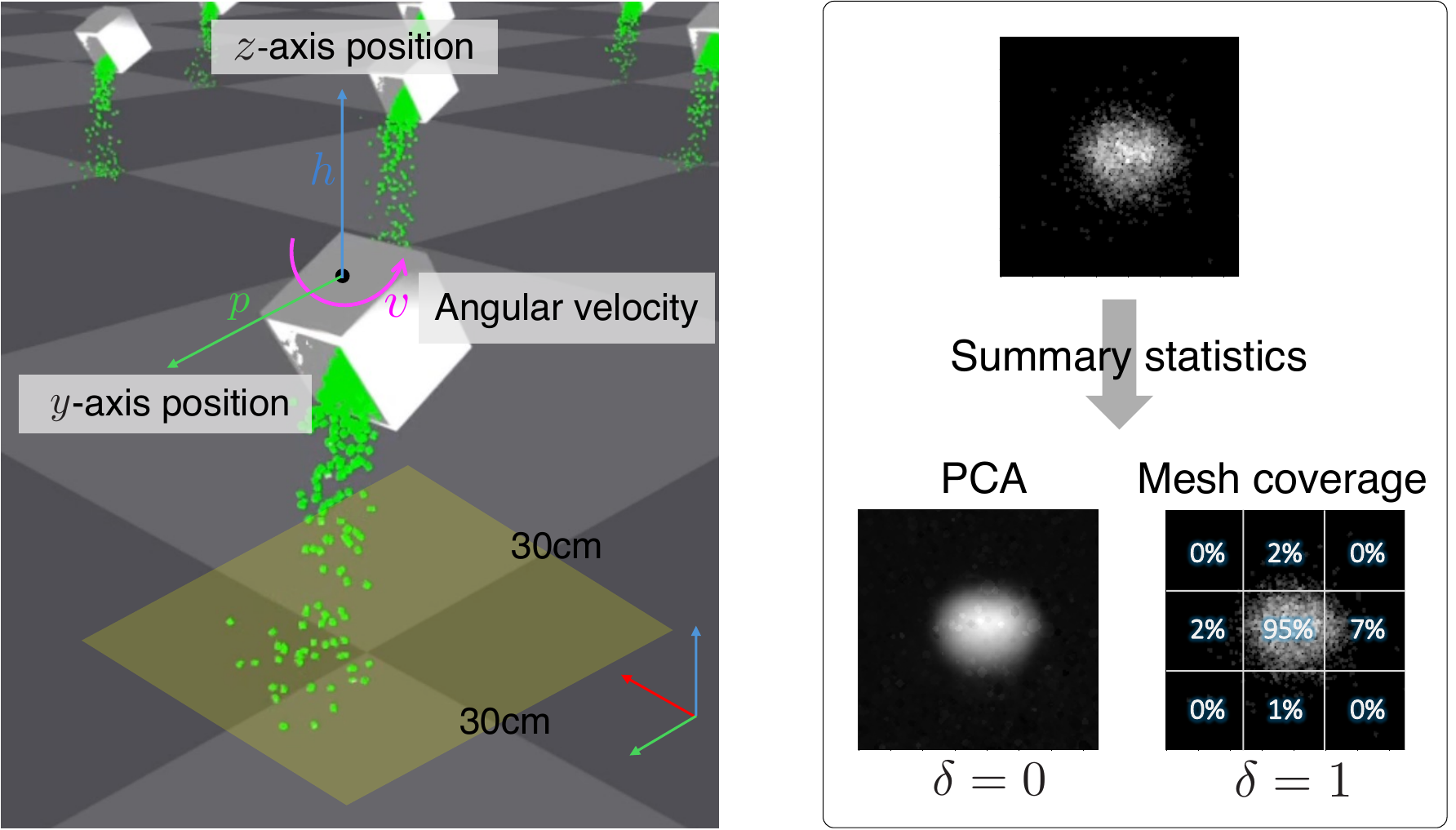}
\caption{Simulation of pouring particles with a bucket. Left: Action variables controlling pouring action. A vector consists of a position on the $z$-axis, a position on the $y$-axis, and angular velocity. A depth image of the 30 cm $\times$ 30 cm area centered at $(x, y) = (0, 0)$ is taken. Right: Types of summary statistics. We also optimize the type of summary statistics between PCA and mesh coverage.}
\label{fig:fig10}
\end{figure}

\subsection{Sim-to-sim experiment with robotics simulator on particle-parameter estimation task}\label{sec:sec5_4}
Next, we verify ASBI with a robotics-purposed black-box simulator equipped with particle models. We perform the particle-parameter estimation task of inferring the properties of cube particles while pouring them with a bucket, as shown in Fig.~\ref{fig:fig10}. We optimize the pouring action by controlling the bucket's position and angular velocity. 

\noindent{}{\bf Simulation Model:} We consider simulations of pouring particle objects with a bucket.  Each side of the bucket is 10 cm. We describe particles using cube objects having sides 0.4 cm $\times$ 0.4 cm $\times$ 0.5 cm and density 1000 kg/m${}^3$. A total of 1000 cubes are contained inside the bucket. After the bucket moves to point $(0, p, h)$, $p$ cm along the $y$-axis and $h$ cm along the $z$-axis, it starts to pour particles while rotating at $v$ rad/sec. After the particles finish pouring, a depth sensor takes a depth image of the 30 cm $\times$ 30 cm area centered at the origin $(0, 0, 0)$. 

\noindent{}{\bf Simulation Parameter:} We estimate 3-dimensional parameter $\btheta  = (\theta_{f}, \theta_{rf}, \theta_{res})$ of the cube object, where $\theta_{f}, \theta_{rf}$, and $\theta_{res}$ represent the coefficient of (static) friction, rolling friction and restitution, respectively. Each particle has the same physical parameter values that range from 0 to 1. The initial prior is assumed as a uniform distribution over this area. 

\begin{figure*}[tb]
    \centering
    \begin{center}
    \includegraphics[width=1.0\hsize]{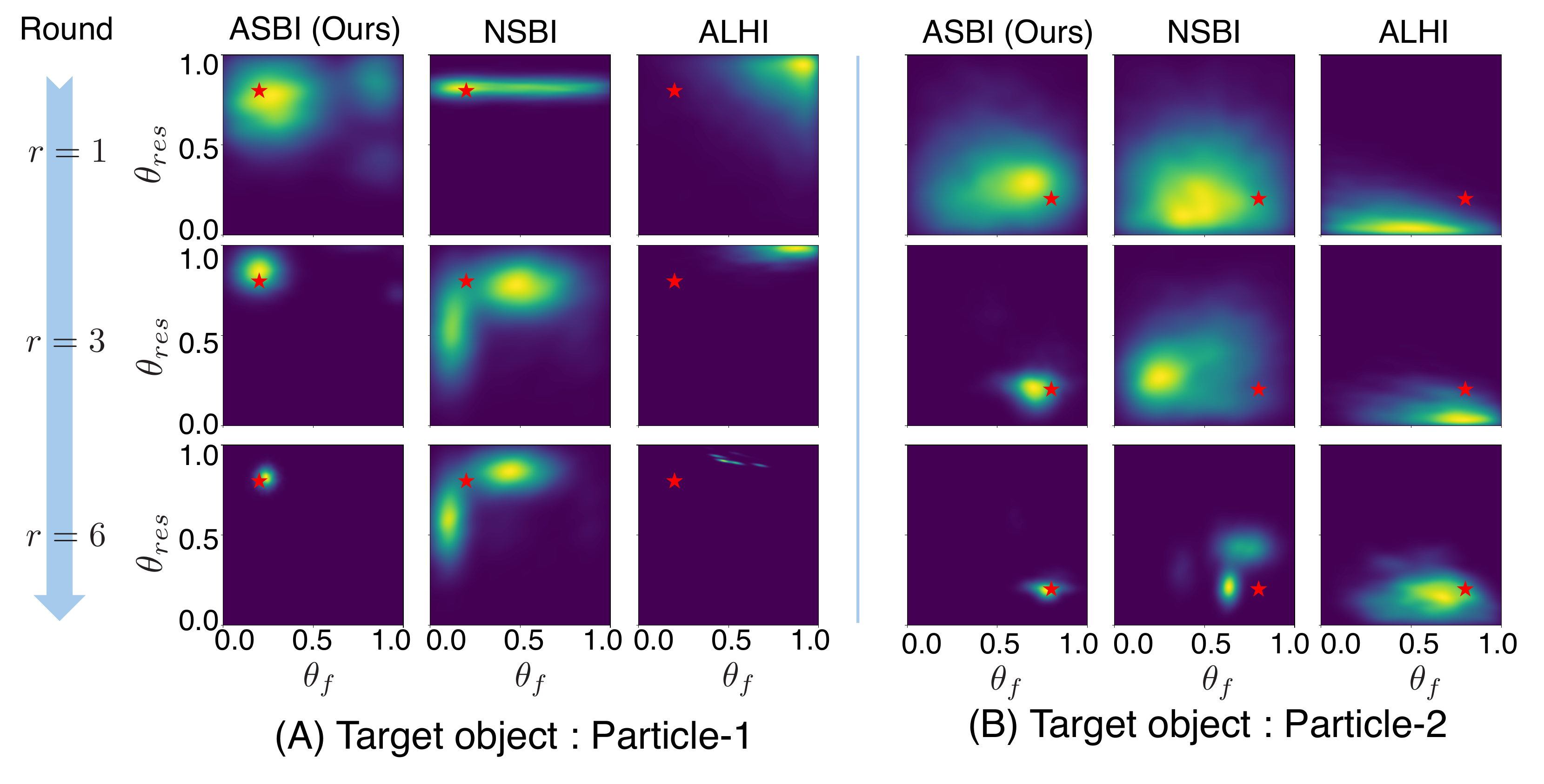}
    \end{center}
\caption{Comparison of the change of posteriors at rounds $r=1, 3, 6$ for (A) Particle-1 and (B) Particle-2 with different methods in the sim-to-sim experiment on the particle-parameter estimation task. For visual simplicity, we draw 2-dimensional marginal distributions of the posterior on friction (x-axis) and restitution (y-axis). True parameters are shown with a red star. } 
\label{fig:fig11}
\end{figure*}

\noindent{}{\bf Action Variables:} Height $h$, $y$-position $p$, and angular velocity $v$ of the bucket are controlled to control the observations. Height $h$ is from $\{20, 80, 140\}$, y-position $p$ from $\{ -30, 0, 30\}$, and $v$ from \{0.5, 2.0, 5.0\}. In addition to the above variables, we also optimize the choice of summary statistics, which determine how raw images are preprocessed. The type of summary statistics, represented as $\delta$, are selected between PCA and the mesh coverage, which are explained next. Including $\delta$, vector $\bxi$ is from the discrete space:

\begin{align}
\begin{split}
    \mathcal{D} = \{ (h, p, v, \delta) | h &\in \{20, 80, 140\}, p \in \{-30, 0, 30\},   
    \\ v &\in \{0.5, 2.0, 5.0\}, \delta \in \{0, 1\} \}.
\end{split}
\end{align}
For the sake of notational convenience, we refer to vector $\bxi = (h, p, v, \delta)$ as an action variable.

\noindent{}{\bf Summary Statistics:} We use two types of summary statistics. The first type is principal component analysis (PCA), which transforms a 22,500-dimensional vector into a 300-dimensional vector. For the second type, we measure how the particles are distributed. We call this summary type the mesh coverage (Fig.~\ref{fig:fig10}). First, we blur a raw image with three blur sizes: 1 pixel, 3 pixels, and 5 pixels. Then, for each blurred image, we divide the image into 12 areas and count the ratio of the covered area versus the entire area, which results in a 36-dimensional vector. Note that the choice of how to preprocess raw data can also be controlled during the estimation process, so it can thus be considered an optimization target.  We denote this choice using a binary variable $\delta$ such that  $\delta=0$ for PCA and $\delta=1$ for the mesh coverage.

\noindent{}{\bf Network Training:} We use two separate networks for the posterior estimator; one is for $\delta=0$, and the other is for $\delta=1$. Both networks have two hidden layers of size (64, 64) and output elements of 5-MoGs. In this experiment, we perform a 6-round parameter estimation. In each round, neural networks are trained in 1000 iterations of batch size 30. The posterior in each round is used as the prior in the next round.

\noindent{}{\bf Utility Calculation:} We calculate the utility \text{Eq.~\ref{eq:finite_sum_utility}} with $M=100$ samples from the training data of neural networks. We calculate this three times and use the average value.

\begin{figure}[tb]
\centering
\includegraphics[width=0.90\columnwidth]{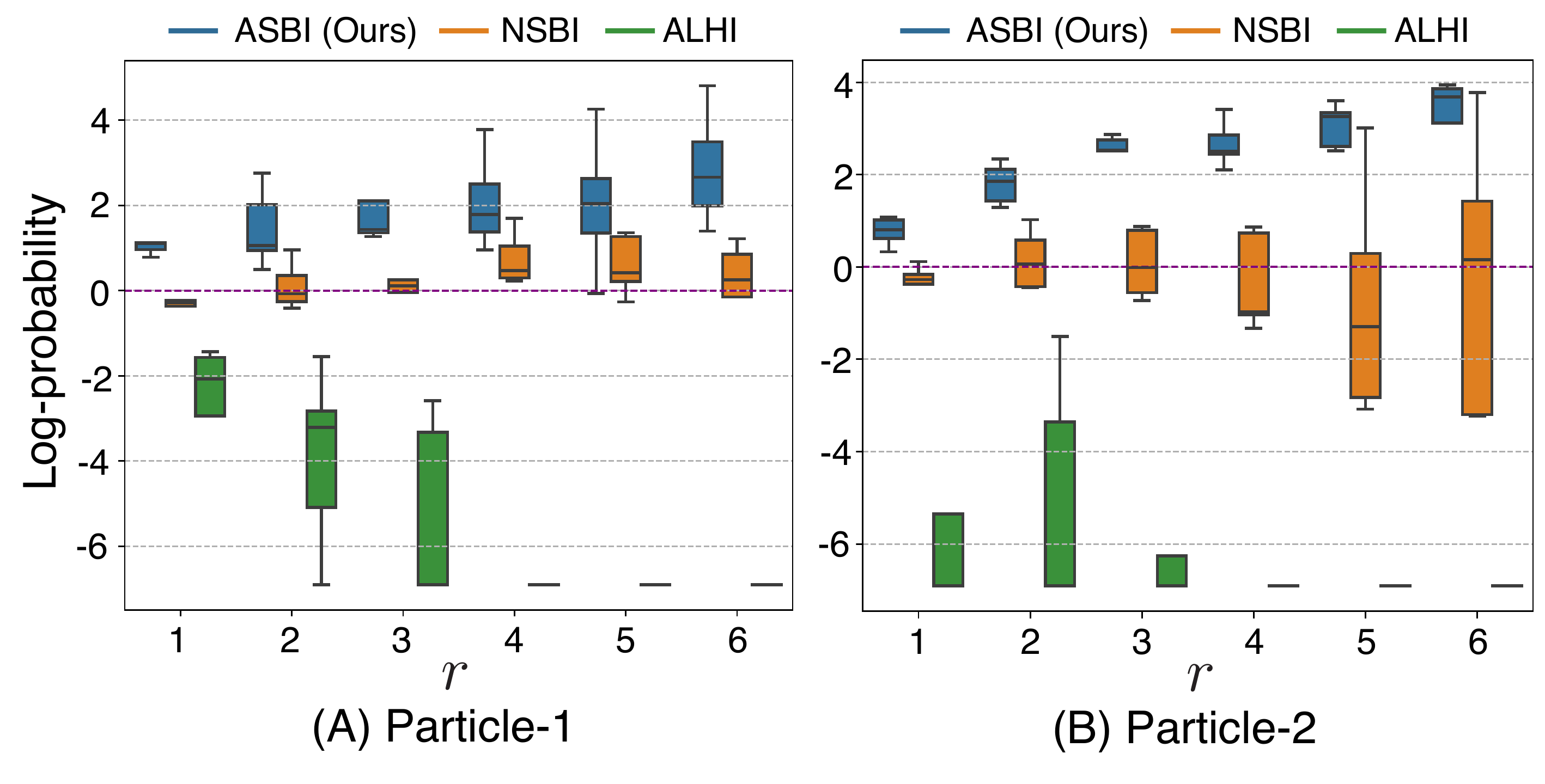}
\caption{Comparison of log probabilities of the true parameters at each round $r$ for the sim-to-sim experiment on the particle-parameter estimation task. Left: Particle-1, Right: Particle-2. The log probability of the initial prior $p(\btheta)=1$ is shown with dotted purple lines. 
 Since ALHI tends to return very low values when proceeding with sequential updates, we replace log-likelihood values smaller than -7 with -7 to draw the above graph. }
\label{fig:fig12}
\end{figure}

\noindent{}{\bf Evaluation:} For quantitative evaluation, we use simulation as a target environment. As the target cube object to estimate, we use two different types of particle objects.
Particle-1 has low friction and high restitution of the value $\btheta^{\text{true}}=(\theta_f^{\text{true}}, \theta_{rf}^{\text{true}}, \theta_{res}^{\text{true}}) = (0.2, 0.2, 0.8)$. Particle-2 has high friction and low restitution  $\btheta^{\text{true}}=(\theta_f^{\text{true}}, \theta_{rf}^{\text{true}}, \theta_{res}^{\text{true}}) = (0.8, 0.8, 0.2)$. For each method, we conduct a 6-round sequential estimation five times.

\noindent{}{\bf Results:} Fig.~\ref{fig:fig11} shows 2-dimensional marginal distributions of the estimated posterior having $\theta_{f}$ as $x$-axis and $\theta_{res}$ as $y$-axis for each target and different method. KDE is used to draw the marginal distributions. For both target parameters, ASBI provides highly peaked distributions around the true parameter, indicating that ASBI provides accurate estimation and successfully reduces uncertainty in prediction. 

Fig.~\ref{fig:fig12} shows the log-probabilities of the true parameters across sequential rounds for the particle-parameter estimation tasks (left: Particle-1, right: Particle-2). Before any updates ($r=0$), the initial prior is a uniform distribution over the area, plotted as the dotted line at $\log p(\btheta^{\text{true}}) = 0.0$. ASBI rapidly improves beyond this uninformed prior after the first round, achieving median log-probabilities above 1.0 for Particle-1 and around 0.8 for Particle-2. As sequential updates progress, ASBI exhibits the fastest posterior convergence, reaching over 2.6 median log-probability for Particle-1 and above 3.6 for Particle-2 by the final round. \par

In contrast, NSBI shows only modest improvements, with medians remaining near zero in later rounds. ALHI collapses below the clipping threshold $-7.0$ as updates progress, indicating numerical instability and failure to refine the posterior effectively. This collapse is likely due to the high-dimensional nature of the observations, which makes learning a surrogate likelihood challenging.  Although ASBI exhibits a wide IQR in round 6 for Particle-1, its minimum log-probability still exceeds the maximum values of both NSBI and ALHI, indicating consistently superior estimates.  For Particle-2, ASBI maintains a tight IQR in the later rounds, highlighting its robustness. \par 

These results reinforce that ASBI's sequential active action selection with action-extended NPE accelerates posterior refinement even under complex, high-dimensional observation settings, achieving both higher accuracy and greater stability compared to baseline methods.

\begin{figure*}[tb]
    \centering
    \begin{center}
    \includegraphics[width=1.0\hsize]{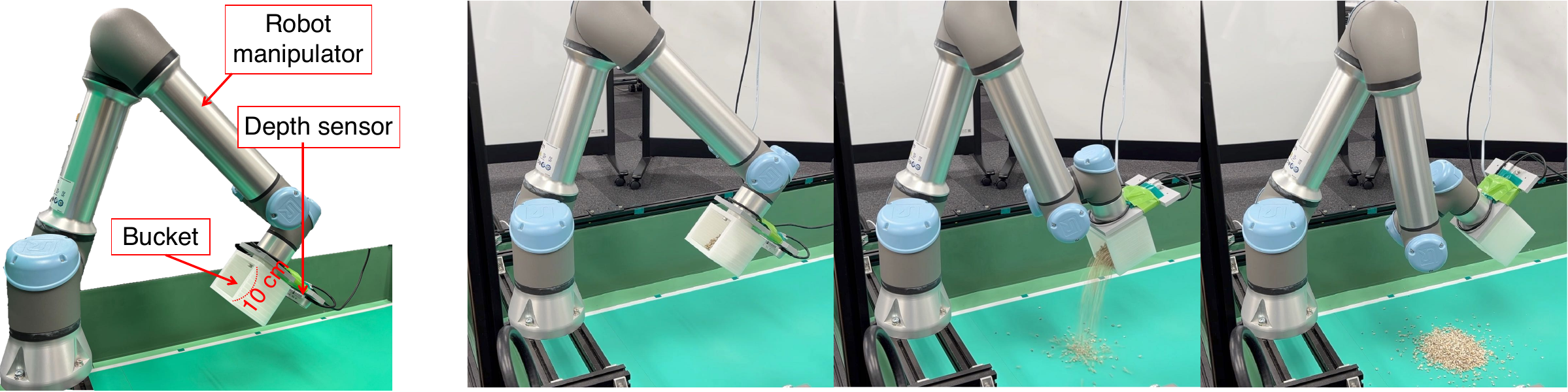}
    \end{center}
\caption{ Real robot setting for the real-to-sim experiment on the particle-parameter estimation task, which estimates the parameters corresponding to beads and gravel. A bucket fabricated by a 3D printer and a depth sensor are attached to the robot manipulator. The robot collects observations by pouring target objects using the bucket.} 
\label{fig:fig13}
\end{figure*}

\begin{figure}[tb]
\centering 
\includegraphics[width=0.6\hsize]{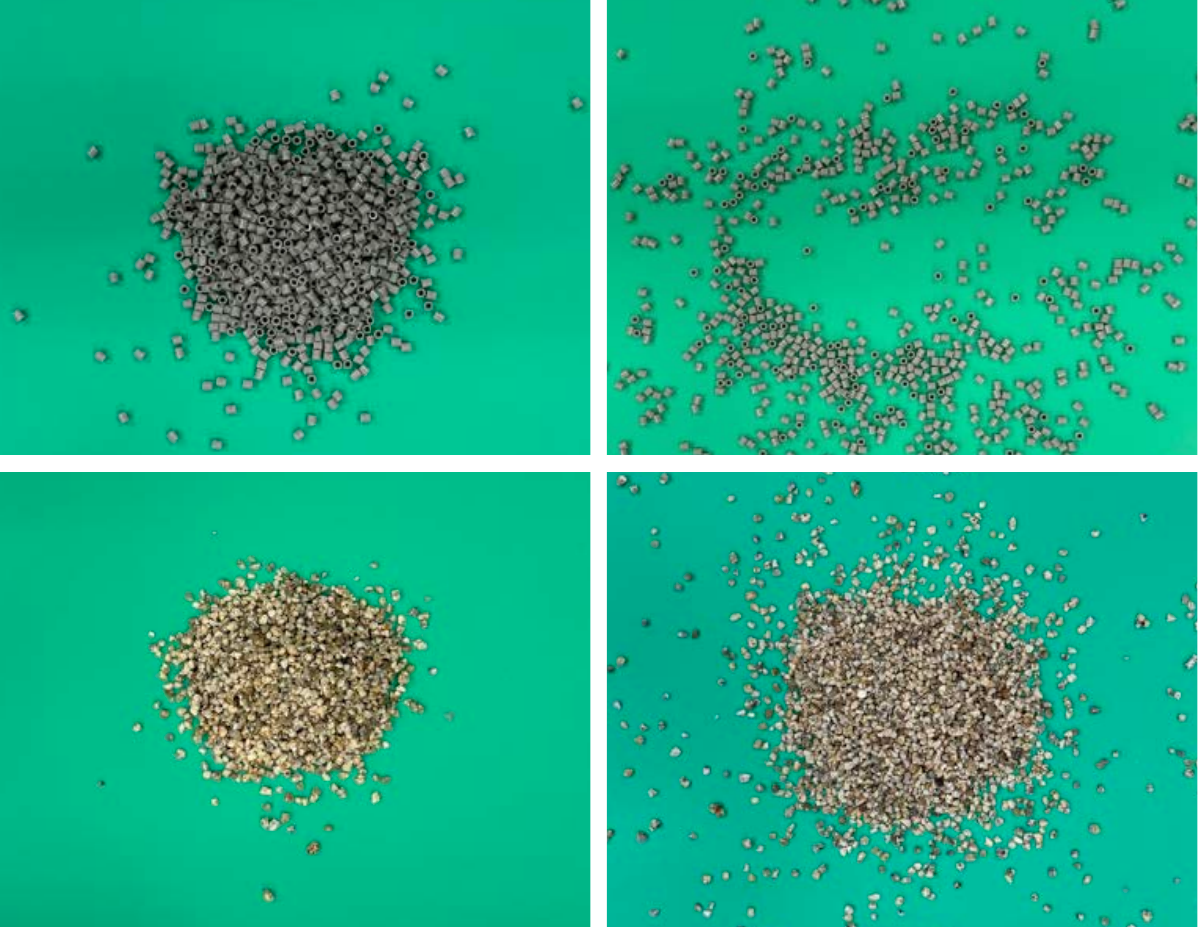}
\caption{Real objects for the real-to-sim experiment. We estimate two different real objects, beads (top) and gravel (bottom), in the real-to-sim robot model. Beads and Gravel have different physical properties and behave differently when dropped with the same action. }
\label{fig:fig14}
\end{figure}

\section{Real Robot Application to Particle-Parameter Estimation Task}\label{sec:sec6}

In this section, we demonstrate the effectiveness of ASBI in real-world usage by adopting a robot manipulator and two real particle objects: beads and gravel.  Similar to the previous sim-to-sim experiment, we perform the particle-parameter estimation task to estimate simulation parameters corresponding to beads and gravel. This task was chosen because replicating particle properties in simulators is crucial for industrial applications, such as digital twin systems in construction operations.

\noindent{}{\bf Experiment Setting:}  In this experiment, we perform a bucket pouring experiment similar to the previous simulation experiment using a robot manipulator (Fig.~\ref{fig:fig13}). A bucket and a depth camera (Intel RealSense D435if) are mounted on a robot arm (UR5e). The bucket has sides of 10 cm and is fabricated by a 3D printer. Two types of target objects are used: beads and gravel (Fig.~\ref{fig:fig14}). We use cylindrical beads with a diameter of 4 mm and a height of 5 mm, and gravel with sides ranging from 2 mm to 6 mm. We measure the volume of 1000 beads and use the same volume of gravel. In simulation, they are represented as 1000 cube objects with sides of 4 mm, 4 mm, and 5 mm. The density of each object is measured physically in advance and then fixed at 400 kg/m${}^3$ for beads and 1430 kg/m${}^3$ for gravel throughout the experiment.

\begin{figure}[tb]
\centering 
\includegraphics[width=0.90\columnwidth]{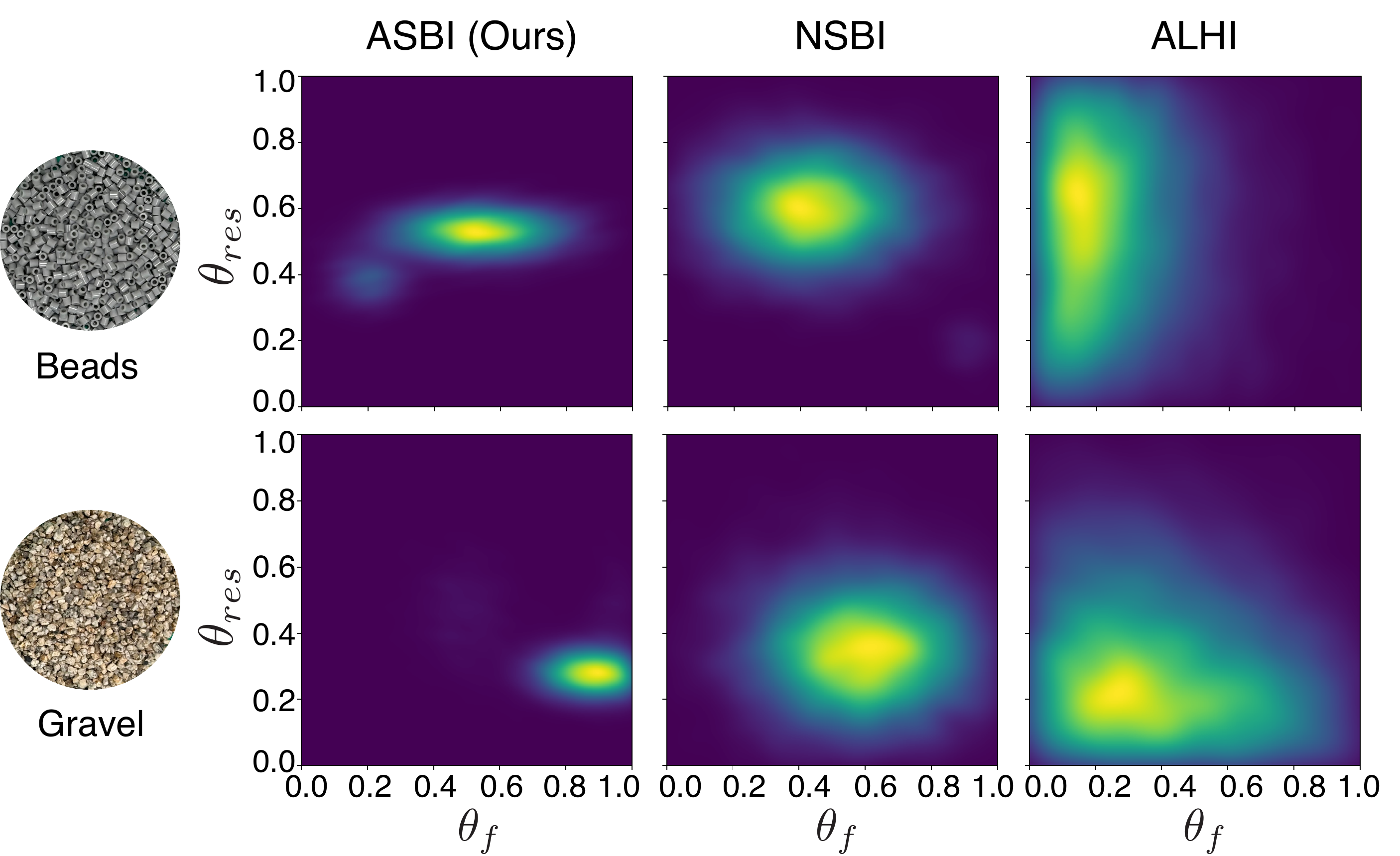}
\caption{Comparison of the final posterior for beads (top) and gravel (bottom) with different methods in the real-to-sim experiment on the particle-parameter estimation task. For visual simplicity, we draw 2-dimensional marginal distributions of the posterior on friction (x-axis) and restitution (y-axis). }
\label{fig:fig15}
\end{figure}

\noindent{}{\bf Simulation Parameter:} Compared to the sim-to-sim experiment, where the ground is identical across all environments, the real-to-sim experiment has nonidentical ground between real-world and simulation environments. Therefore, we also estimate the parameters of the ground since the results differ depending on the property of the ground. In this experiment, we estimate 8-dimensional parameter $\btheta=(\theta_{gf}, \theta_{gdf}, \theta_{gres}, \theta_{f}, \theta_{rf}, \theta_{res}, \theta_{ld}, \theta_{ad})$, where $\theta_{gf}$, $\theta_{gdf}$, and $\theta_{gres}$ represent the coefficient of static friction, dynamic friction, and restitution of the ground, respectively; $\theta_{f}, \theta_{rf}$, and $\theta_{res}$ denote the coefficients of static friction, rolling friction, and restitution for the cubes, consistent with the previous experiment; and $\theta_{ld}, \theta_{ad}$ correspond to the linear and angular damping coefficients of the cube. We assume that all parameters range from 0 to 1 and that the initial prior is a uniform distribution.

\noindent{}{\bf Action Variables:} Considering the physical limitations of the real robot, height $h$ is from \{15, 35, 55\} and velocity $v$ is from \{0.2, 0.9, 1.6\}. The $y$-position $p$ is the same as in the previous experiment. Also, the same as in the previous sim-to-sim experiment, we include the choice of summary statistics $\delta$. Including $\delta$, vector $\bxi$ is from the discrete space 
\begin{align}
\begin{split}
    \mathcal{D} = \{ (h, p, v, \delta) | h &\in \{15, 35, 55\}, p \in \{-30, 0, 30\},   
    \\ v &\in \{0.2, 0.9, 1.6\}, \delta \in \{0, 1\} \}.
\end{split}
\end{align}
For the sake of notational convenience, we refer to vector $\bxi = (h, p, v, \delta)$ as an action variable.

\noindent{}{\bf Summary Statistics:} We consider two summary statistics, PCA ($\delta=0$) and the mesh coverage ($\delta=1$), which are the same as in the previous experiment. 

\noindent{}{\bf Network Training:} The structure of neural networks is the same as that in the previous experiment. In this experiment, we perform a 4-round parameter estimation. In each round, neural networks are trained in 1000 iterations of batch size 30. The posterior in each round is used as the prior in the next round.

\noindent{}{\bf Utility Calculation:} We calculate the utility Eq.~\ref{eq:finite_sum_utility} with $M=100$ samples from the training dataset of neural networks. We calculate this three times and use the average value.

\begin{figure}[tb]
\centering 
\includegraphics[width=0.90\columnwidth]{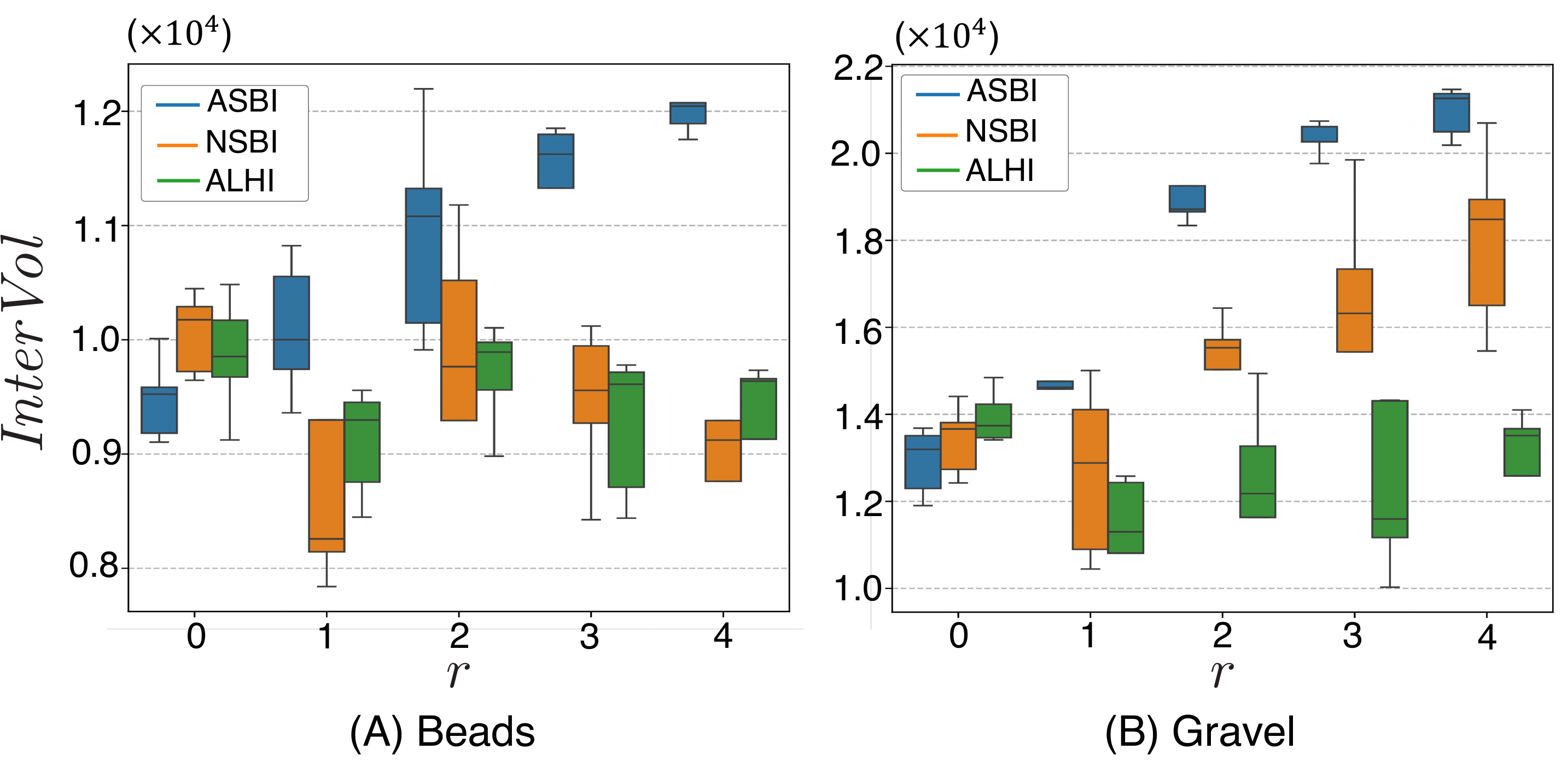}
\caption{Comparison of intersection volume across rounds in the real-to-sim experiment on the particle-parameter estimation task, with beads (left) and gravel (right). Estimation using ASBI achieves consistently higher intersection volumes and more closely replicates the real-world behavior compared to the other methods. }
\label{fig:fig16}
\end{figure}

\noindent{}{\bf Evaluation:} Since the true parameters of real target objects are unobservable, we measure the similarity between the simulated and the true depth images. For depth images $X_1$ and $X_2$, we define the intersection volume of two images as follows:

\begin{equation}
    G(X_1, X_2) = \sum_{a \in A} \mathrm{min}(X_1(a), X_2(a))
\end{equation}
where $a$ is a pixel position of intersection area $A$, and $X(a)$ means the value of depth image $X$ at pixel $a$. This measures the intersected area of $X_1$ and $X_2$ and represents how similar the two depth images are, since the entire volume of all objects is fixed. After each round of estimation, we sample $N=100$ parameters from the posterior and perform a simulation with middle position and middle velocity $(h, v, p)=(35, 0.9, 0)$ for each parameter. Then, we measure the average intersection volume between the true depth image and the  simulated image:

\begin{equation}\label{eq:exp3_intervol}
\textit{InterVol} = \frac{1}{N} \sum^{N}_{n=1} G(X^{\text{real}}, X^{\text{sim}}_n)
\end{equation}
where $X^{\text{real}}$ is the true depth image of real objects and $X^{\text{sim}}$ is a simulated depth image of a sampled parameter. This metric evaluates how well the calibrated simulator matches the real world, with higher values indicating better estimation.

\noindent{}{\bf Results:}  Fig.~\ref{fig:fig15} shows marginal posteriors for object friction and restitution obtained in
the final rounds. ASBI successfully narrows down the posterior compared to the other methods. In addition, we can confirm that beads have lower friction and higher restitution than gravel, which is consistent with our intuition. 

Fig.~\ref{fig:fig16} shows the intersection volume ($InterVol$) for the real robotic experiments on Beads (left) and Gravel (right).  We conduct five experiments for each object and method. ASBI consistently achieves the largest increase in intersection volume across rounds, while ALHI and NSBI exhibit limited or unstable improvements. 

For the Beads, ASBI increases the median intersection volume from approximately $0.95 \times 10^4$ in round 0 to $1.20 \times 10^4$ in round 4, corresponding to an improvement of about $26\%$. In contrast, ALHI and NSBI remain close to their initial performance without a clear upward trend. At round 4, ALHI and NSBI yield approximately $20\%$ and $24\%$ lower median values than ASBI, respectively, highlighting ASBI's superior calibration ability.  

For the Gravel experiment, ASBI achieves a larger improvement, increasing the median intersection volume from $1.32 \times 10^4$ to $2.13 \times 10^4$, which corresponds to an improvement of roughly $61 \%$, whereas ALHI and NSBI fail to achieve comparable gains. At round 4, ALHI and NSBI yield approximately $37 \%$ and $13 \%$ lower median values than ASBI, respectively, further confirming the effectiveness of ASBI in achieving more accurate sim-to-real calibration. 

Moreover, IQR for ASBI remains narrower in later rounds, indicating more stable and consistent performance, while ALHI and NSBI show wider IQRs, reflecting higher variability. These results confirm that ASBI's sequential active action selection with action-extended NPE effectively improves real-world reproduction performance with fewer rounds, achieving both higher median intersection volumes and greater stability, even for materials with complex and uncertain physical properties.

\section{Discussion}\label{sec:sec7}

{\bf Assumption on Model Discrepancy and Sensor Noise:} In this paper, we assume that the simulator and real-world share an identical dynamics model, neglecting potential discrepancies such as sensor noise and modeling errors. While this assumption simplifies the estimation process, it is often unrealistic because simulations inherently contain approximation errors. Such discrepancies can bias parameter estimation and reduce its applicability to real-world scenarios. 

To address these challenges, recent work on neural posterior estimation (NPE) under model mismatch, such as Robust Posterior Estimation (RNPE)~\cite{Ward2022-fq}, proposes strategies for mitigating bias caused by imperfect simulators. Since ASBI builds upon NPE as a core component, robustness techniques developed for NPE can be used in the same manner. In addition, recent advances in robust visual tracking~\cite{Tian2025-dd, Xu2025-xd} highlight how noisy or adversarially perturbed observations can negatively impact perception and decision-making. These studies also propose effective strategies to mitigate such effects, which could be applied to ASBI to enhance its robustness against sensor noise. 

In future work, we plan to explicitly account for modeling errors and improve robustness under real-world conditions by integrating methods such as robust neural posterior estimators or sensor-noise denoising methods, thereby bridging the gap between simulation and reality and improving robustness against noisy or adversarial observations.

{\bf Scalability in High-Dimensional Parameter and Action Spaces:} Scalability is a key consideration for extending ASBI to complex robotic systems with high-dimensional parameter or action spaces. In parameter estimation, higher-dimensional settings generally require more simulation data to model complex posterior distributions. ASBI addresses this through its sequential design: neural network training is divided into multiple rounds, and in each round, the sampling distribution is updated using the latest prior. This iterative refinement improves sample efficiency by concentrating on high-probability regions of the parameter space. Moreover, in many robotic scenarios, the observation space is much higher-dimensional than the parameter space, making posterior-based methods more practical and robust than surrogate likelihood-based approaches.

For action selection, ASBI currently employs a greedy search strategy. While effective in relatively small action spaces, its computational cost grows linearly with the number of candidate actions, which limits scalability in more complex or continuous action domains. To mitigate this, more efficient optimization strategies can be adopted. Bayesian optimization~\cite{Kleinegesse2021-zj, Shahriari2016-fy, Nguyen2019-ww} offers a sample-efficient method for navigating large or continuous action spaces, while gradient-based optimization with differentiable posterior estimators~\cite{Rumelhart1986-jc, Hecht1989-ox} can further accelerate the process. These alternatives reduce computational overhead while preserving ASBI's adaptive and informative action selection, making the framework more suitable for large-scale robotic tasks.

{\bf Difference from End-to-End Recognition Methods:} Although ASBI and end-to-end recognition methods such as YOLO-based object detection~\cite{Tao2024-ce, Tao2025-im} and multi-scale residual networks~\cite{Sun2025-dp} all employ neural networks to map observations, such as images, to task-relevant outputs, their goals and paradigms are fundamentally different. Recognition networks aim to learn a universal, static feature extractor that processes arbitrary inputs for perception tasks. In contrast, ASBI focuses on a specific target environment with sequential inference, where the posterior over parameters is iteratively updated as new observations are collected. This distinction is crucial in robotics, where robots can collect new observations from the target environment, and the collecting actions directly influence what information can be obtained. The primary goal of ASBI is to minimize the number of required real-world observations during online exploration by adaptively selecting robot actions that maximize information gain, thereby enabling efficient information acquisition in costly, real-world robotic executions. 

{\bf Hyperparameter Selection and Practical Considerations:} In this study, the hyperparameters of ASBI, such as the network architecture of the neural posterior estimator and the number of training epochs, were manually selected without extensive fine-tuning. Our primary focus was to demonstrate the effectiveness of sequential active action selection in conjunction with the posterior estimator rather than to optimize network configurations. In practice, SNPE methods often adopt early stopping based on validation performance, where training is halted if no improvement is observed on a held-out validation set for a fixed number of iterations. This strategy helps prevent overfitting and stabilizes training without requiring exhaustive hyperparameter searches. The same approach can be applied to ASBI in the same way as SNPE. While such fine-tuning could further improve stability and accuracy, it is beyond the scope of this work and can be considered for practical deployment or future extensions.

{\bf Applicable Domain:} The proposed method is particularly suitable for robotic tasks that allow online interaction with the environment, where robots can actively collect data and sequentially update their beliefs over unknown parameters. In particular, it is effective for calibration in scenarios with uncertain and varying dynamics, where ad hoc parameter estimation is required for reliable operation. For example, the method is well-suited for tactile-based manipulation tasks~\cite{Li2020-ia, Kaboli2016-ic, Dutta2023-qq}, where knowledge of object properties significantly enhances task performance. Furthermore, it generalizes to a wide range of robotic domains, including terrain-adaptive mobile robotics~\cite{Margolis2023-ts, Puentes2024-ot}, underwater exploration~\cite{Huajun2020-yz, Rasul2024-jl}, and biomedical robotics~\cite{Vargas2024-bl, Cornejo2024-jb}, where robots can actively collect data from the environment, and the initial calibration step is critical for safety and effectiveness. Across these domains, the ability to actively reduce uncertainty through interaction makes the proposed method a powerful tool for improving task performance and robustness. 

{\bf Computational Efficiency and Real-Time Feasibility:} ASBI consists of three main computational components: simulation, neural network training, and information gain calculation. Among these, simulation time is the dominant overhead and heavily depends on the complexity of the task. In our runtime analysis (Section~\ref{sec:sec5_3}), the total time per iteration was approximately 35 seconds, indicating that ASBI can operate in near real-time for relatively simple tasks. However, in more complex scenarios, the simulation step can be significantly more time-consuming. In some cases, it may exceed an hour per iteration, which makes real-time execution infeasible without further optimization. One promising approach to address this limitation is to pre-collect a large offline dataset of simulations and perform sample reweighting based on the updated prior distribution during inference. Although this strategy requires substantial computational resources upfront, it removes the need for online simulation and allows ASBI to achieve near real-time inference at deployment. In addition, techniques such as parallel computation of information gain and early stopping during training can further improve runtime efficiency.

{\bf Robustness of NPE training:} It is important to note that ASBI relies on the Neural Posterior Estimator (NPE) to approximate the posterior $p(\btheta | \bx, \bxi)$, and thus the quality of the action selection depends on how well the NPE is trained. In this work, we assume that the NPE is sufficiently trained, and we did not explicitly analyze the sensitivity of ASBI to posterior miscalibration or approximation errors. In practice, poorly trained or underfitted NPE models could lead to inaccurate information gain estimates, which may result in suboptimal action selection. This limitation could be mitigated by integrating techniques for NPE considering model mismatch and robustness, such as Robust Neural Posterior Estimation (RNPE)~\cite{Ward2022-fq}. A detailed investigation of ASBI robustness to posterior approximation errors remains an important direction for future work.

{\bf Extension Beyond Simulator Calibration:} While ASBI is primarily developed for simulator calibration, the framework can be naturally extended to black-box policy learning. Previous studies such as BayesSim~\cite{Ramos2019-oy} and NPDR~\cite{Muratore2022-cy} have demonstrated that refining simulation parameters leads to improved policy accuracy in reinforcement learning. Similarly, ASBI could be employed to iteratively update simulation parameters during policy learning, thereby improving sim-to-real transfer accuracy through more realistic simulators. However, extending ASBI to policy learning presents additional challenges, particularly in balancing the learning between the exploration for information gain and optimization for task rewards.

{\bf Exploration-Exploitation Trade-off:} In the current ASBI framework, the optimal action is selected using a greedy strategy, where the expected information gain is evaluated across all candidate actions, and the action with the highest score is chosen. While this approach effectively maximizes immediate information gain, it does not explicitly account for the exploration-exploitation trade-off. As a result, the method may overfit to regions that initially appear informative, potentially overlooking other areas that could contribute to improved long-term performance.

Future work could address this limitation by integrating ASBI with broader sequential decision-making strategies. For example, bandit algorithms can explicitly manage the exploration-exploitation trade-off over multiple rounds, Bayesian optimization provides a sample-efficient way to navigate large or continuous action spaces, and reinforcement learning can optimize long-term cumulative information gain rather than focusing solely on myopic rewards. Such integration would enhance scalability, improve the exploration-exploitation balance, and increase adaptability in more complex robotic tasks.

\section{Conclusion}\label{sec:sec8}

  In this paper, we introduced a practical parameter-tuning framework for black-box robotics simulators that performs active action selection, enabling accurate parameter estimation with only a few real executions. We demonstrated that leveraging robot actions that maximize information gain effectively reduces uncertainty in the estimated posterior of parameters. Unlike previous methods that rely on training surrogate likelihood models to calculate the posterior and information gain, our approach directly learns the posterior from simulation data using modern likelihood-free inference techniques. This direct approach is advantageous because the parameter space typically has a much lower dimensionality compared to the observation space. Our approach avoids the need to learn surrogate likelihood models and compute the posterior analytically using the Bayes' rule, making our method both simpler and more efficient. 
  
  Through three sim-to-sim experiments, we verified that the proposed method achieves higher accuracy with a small number of observations compared to random action selection or likelihood-based approaches. Furthermore, we illustrated our method's potential in real-world application through a real-to-sim experiment on the particle-parameter estimation task. In future work, we plan to extend our framework to more complex real-world tasks, such as estimating various soil properties using earthwork vehicles and validating improvements in manipulation accuracy achieved through simulator tuning.

\backmatter

\bmhead{Acknowledgements}
This work is supported by JST [Moonshot Research and Development], Grant Number [JPMJMS2032].

\bmhead{Authors contributions} Gahee Kim: Conceptualization, Methodology, Software, Experiment, Analysis, Visualization, Writing. Takamitsu Matsubara: Conceptualization, Writing - Review \& Editing, Supervision, Project administration. 

\bmhead{Data Availability}
No datasets were generated or analyzed during the current study.

\section*{Declarations}

\bmhead{Competing Interests}  The authors have no competing interests to declare.

\bmhead{Ethical and informed consent for data used} This article does not contain any studies with human participants or animals.

\bibliography{sn-bibliography}

\begin{thebibliography}{52}
\providecommand{\natexlab}[1]{#1}
\providecommand{\url}[1]{{#1}}
\providecommand{\urlprefix}{URL }
\providecommand{\doi}[1]{\url{https://doi.org/#1}}
\providecommand{\eprint}[2][]{\url{#2}}
 \bibcommenthead

\bibitem[{Cranmer et~al.(2020)Cranmer, Brehmer, and Louppe}]{Cranmer2020-cc}
Cranmer K, Brehmer J, Louppe G (2020) The frontier of simulation-based inference. Proc Natl Acad Sci U S A 117(48):30055--30062. \doi{10.1073/pnas.1912789117}

\bibitem[{Pina-Otey et~al.(2020)Pina-Otey, Sánchez, Gaitan, and Lux}]{Pina2020-xv}
Pina-Otey S, Sánchez F, Gaitan V, et~al (2020) Likelihood-free inference of experimental neutrino oscillations using neural spline flows. Phys Rev D 101:113001. \doi{10.1103/PhysRevD.101.113001}

\bibitem[{Lueckmann et~al.(2017)Lueckmann, Gonçalves, Bassetto, Öcal, Nonnenmacher, and Macke}]{Lueckmann2017-sa}
Lueckmann JM, Gonçalves PJ, Bassetto G, et~al (2017) Flexible statistical inference for mechanistic models of neural dynamics. In: Advances in Neural Information Processing Systems, p 1289–1299

\bibitem[{{Vasist, Malavika} et~al.(2023){Vasist, Malavika}, {Rozet, François}, {Absil, Olivier}, {Mollière, Paul}, {Nasedkin, Evert}, and {Louppe, Gilles}}]{Vasist2023-cv}
{Vasist, Malavika}, {Rozet, François}, {Absil, Olivier}, et~al (2023) Neural posterior estimation for exoplanetary atmospheric retrieval. Astron Astrophys 672:A147. \doi{10.1051/0004-6361/202245263}

\bibitem[{Dax et~al.(2021)Dax, Green, Gair, Macke, Buonanno, and Schölkopf}]{Dax2021-rt}
Dax M, Green SR, Gair J, et~al (2021) Real-time gravitational wave science with neural posterior estimation. Phys Rev Lett 127:241103. \doi{10.1103/PhysRevLett.127.241103}

\bibitem[{Khullar et~al.(2022)Khullar, Nord, Ćiprijanović, Poh, and Xu}]{Khullar2022-kv}
Khullar G, Nord B, Ćiprijanović A, et~al (2022) {DIGS}: deep inference of galaxy spectra with neural posterior estimation. Mach Learn: Sci Technol 3(4):04LT04. \doi{10.1088/2632-2153/ac98f4}

\bibitem[{Lips et~al.(2024)Lips, De~Gusseme, and Wyffels}]{Lips2024-bc}
Lips T, De~Gusseme VL, Wyffels F (2024) Learning keypoints for robotic cloth manipulation using synthetic data. IEEE Robot Autom Lett 9(7):6528--6535. \doi{10.1109/LRA.2024.3405335}

\bibitem[{Sundaresan et~al.(2020)Sundaresan, Grannen, Thananjeyan, Balakrishna, Laskey, Stone, Gonzalez, and Goldberg}]{Sundaresan2020-cv}
Sundaresan P, Grannen J, Thananjeyan B, et~al (2020) Learning rope manipulation policies using dense object descriptors trained on synthetic depth data. In: 2020 IEEE International Conference on Robotics and Automation (ICRA), pp 9411--9418, \doi{10.1109/ICRA40945.2020.9197121}

\bibitem[{Mou et~al.(2022)Mou, Wang, and Wu}]{Mou2022-py}
Mou F, Wang B, Wu D (2022) Learning-based cable coupling effect modeling for robotic manipulation of heavy industrial cables. Sci Rep 12(1):6036. \doi{10.1038/s41598-022-09643-6}

\bibitem[{Kadokawa et~al.(2023)Kadokawa, Hamaya, and Tanaka}]{Kadokawa2023-lc}
Kadokawa Y, Hamaya M, Tanaka K (2023) Learning robotic powder weighing from simulation for laboratory automation. In: 2023 IEEE/RSJ International Conference on Intelligent Robots and Systems (IROS), pp 2932--2939, \doi{10.1109/IROS55552.2023.10342463}

\bibitem[{Egli et~al.(2022)Egli, Gaschen, Kerscher, Jud, and Hutter}]{Egli2022-ca}
Egli P, Gaschen D, Kerscher S, et~al (2022) Soil-adaptive excavation using reinforcement learning. IEEE Robot Autom Lett 7(4):9778--9785. \doi{10.1109/LRA.2022.3189834}

\bibitem[{Mittal et~al.(2023)Mittal, Yu, Yu, Liu, Rudin, Hoeller, Yuan, Singh, Guo, Mazhar, Mandlekar, Babich, State, Hutter, and Garg}]{Mittal2023-yd}
Mittal M, Yu C, Yu Q, et~al (2023) Orbit: A unified simulation framework for interactive robot learning environments. IEEE Robot Autom Lett 8(6):3740--3747. \doi{10.1109/LRA.2023.3270034}

\bibitem[{{Algoryx Simulations}(2025)}]{algoryx-2025}
{Algoryx Simulations} (2025) {AGX} dynamics. https://www.algoryx.se/agx-dynamics, {Accessed: 06 March 2025}

\bibitem[{Lindley(1956)}]{Lindley1956-ge}
Lindley DV (1956) On a measure of the information provided by an experiment. Ann Math Stat 27(4):986--1005. \doi{10.1214/aoms/1177728069}

\bibitem[{Papamakarios and Murray(2016)}]{Papamakarios2016-ki}
Papamakarios G, Murray I (2016) Fast $\epsilon$-free inference of simulation models with bayesian conditional density estimation. In: Advances in Neural Information Processing Systems, pp 1028--1036

\bibitem[{Greenberg et~al.(2019)Greenberg, Nonnenmacher, and Macke}]{Greenberg2019-on}
Greenberg D, Nonnenmacher M, Macke J (2019) Automatic posterior transformation for likelihood-free inference. In: Proceedings of the 36th International Conference on Machine Learning, pp 2404--2414

\bibitem[{Kleinegesse et~al.(2021)Kleinegesse, Drovandi, and Gutmann}]{Kleinegesse2021-zj}
Kleinegesse S, Drovandi C, Gutmann MU (2021) Sequential bayesian experimental design for implicit models via mutual information. Bayesian Anal 16(3):773--802. \doi{10.1214/20-BA1225}

\bibitem[{Pritchard et~al.(1999)Pritchard, Seielstad, Perez-Lezaun, and Feldman}]{Pritchard1999-th}
Pritchard JK, Seielstad MT, Perez-Lezaun A, et~al (1999) Population growth of human {Y} chromosomes: a study of {Y} chromosome microsatellites. Mol Biol Evol 16(12):1791--1798. \doi{10.1093/oxfordjournals.molbev.a026091}

\bibitem[{Beaumont et~al.(2002)Beaumont, Zhang, and Balding}]{Beaumont2002-sj}
Beaumont MA, Zhang W, Balding DJ (2002) Approximate bayesian computation in population genetics. Genetics 162(4):2025--2035. \doi{10.1093/genetics/162.4.2025}

\bibitem[{Marjoram et~al.(2003)Marjoram, Molitor, Plagnol, and Tavaré}]{Marjoram2003-th}
Marjoram P, Molitor J, Plagnol V, et~al (2003) Markov chain monte carlo without likelihoods. Proc Natl Acad Sci U S A 100(26):15324--15328. \doi{10.1073/pnas.0306899100}

\bibitem[{Bonassi and West(2015)}]{Bonassi2015-ly}
Bonassi FV, West M (2015) Sequential monte carlo with adaptive weights for approximate bayesian computation. Bayesian Anal 10(1):171--187. \doi{10.1214/14-BA891}

\bibitem[{Possas et~al.(2020)Possas, Barcelos, Oliveira, Fox, and Ramos}]{Possas2020-yh}
Possas R, Barcelos L, Oliveira R, et~al (2020) Online {BayesSim} for combined simulator parameter inference and policy improvement. In: 2020 IEEE/RSJ International Conference on Intelligent Robots and Systems (IROS), pp 5445--5452, \doi{10.1109/IROS45743.2020.9341401}

\bibitem[{Muratore et~al.(2022)Muratore, Gruner, Wiese, Belousov, Gienger, and Peters}]{Muratore2022-cy}
Muratore F, Gruner T, Wiese F, et~al (2022) Neural posterior domain randomization. In: Proceedings of the 5th Conference on Robot Learning, pp 1532--1542

\bibitem[{Myung et~al.(2013)Myung, Cavagnaro, and Pitt}]{Myung2013-wh}
Myung JI, Cavagnaro DR, Pitt MA (2013) A tutorial on adaptive design optimization. J Math Psychol 57(3):53--67. \doi{10.1016/j.jmp.2013.05.005}

\bibitem[{Dushenko et~al.(2020)Dushenko, Ambal, and McMichael}]{Dushenko2020-xd}
Dushenko S, Ambal K, McMichael RD (2020) Sequential bayesian experiment design for optically detected magnetic resonance of nitrogen-vacancy centers. Phys Rev Appl 14:054036. \doi{10.1103/PhysRevApplied.14.054036}

\bibitem[{Rainforth et~al.(2024)Rainforth, Foster, Ivanova, and Smith}]{Rainforth2024-kn}
Rainforth T, Foster A, Ivanova DR, et~al (2024) Modern bayesian experimental design. Statistical Science 39(1):100--114. \doi{10.1214/23-STS915}

\bibitem[{Ryan et~al.(2016)Ryan, Drovandi, McGree, and Pettitt}]{Ryan2016-mo}
Ryan EG, Drovandi CC, McGree JM, et~al (2016) A review of modern computational algorithms for bayesian optimal design. Int Stat Rev 84(1):128--154. \doi{10.1111/insr.12107}

\bibitem[{Saal et~al.(2010)Saal, Ting, and Vijayakumar}]{Saal2010-uf}
Saal HP, Ting JA, Vijayakumar S (2010) Active estimation of object dynamics parameters with tactile sensors. In: 2010 IEEE/RSJ International Conference on Intelligent Robots and Systems (IROS), pp 916--921, \doi{10.1109/IROS.2010.5649191}

\bibitem[{Cooper et~al.(2021)Cooper, McGree, Molloy, and Ford}]{Cooper2021-qg}
Cooper M, McGree J, Molloy TL, et~al (2021) Bayesian experimental design with application to dynamical vehicle models. IEEE Trans Robot 37(5):1844--1851. \doi{10.1109/TRO.2021.3063977}

\bibitem[{Dutta et~al.(2023)Dutta, Burdet, and Kaboli}]{Dutta2023-qq}
Dutta A, Burdet E, Kaboli M (2023) Push to know! - visuo-tactile based active object parameter inference with dual differentiable filtering. In: 2023 IEEE/RSJ International Conference on Intelligent Robots and Systems (IROS), pp 3137--3144, \doi{10.1109/IROS55552.2023.10341832}

\bibitem[{Margolis et~al.(2023)Margolis, Fu, Ji, and Agrawal}]{Margolis2023-ts}
Margolis GB, Fu X, Ji Y, et~al (2023) Learning to see physical properties with active sensing motor policies. In: Proceedings of the 7th Conference on Robot Learning, pp 2537--2548

\bibitem[{Memmel et~al.(2024)Memmel, Wagenmaker, Zhu, Yin, Fox, and Gupta}]{Memmel2024-kk}
Memmel M, Wagenmaker A, Zhu C, et~al (2024) {ASID}: Active exploration for system identification in robotic manipulation. arXiv preprint arXiv:2404.12308. \urlprefix\url{https://arxiv.org/abs/2404.12308}

\bibitem[{Thomas et~al.(2022)Thomas, Dutta, Corander, Kaski, and Gutmann}]{Thomas2022-eu}
Thomas O, Dutta R, Corander J, et~al (2022) Likelihood-free inference by ratio estimation. Bayesian Anal 17(1):1--31. \doi{10.1214/20-BA1238}

\bibitem[{Ramos et~al.(2019)Ramos, Possas, and Fox}]{Ramos2019-oy}
Ramos F, Possas R, Fox D (2019) {BayesSim}: Adaptive domain randomization via probabilistic inference for robotics simulators. In: Proceedings of Robotics: Science and Systems XV, \doi{10.15607/rss.2019.xv.029}

\bibitem[{Papamakarios et~al.(2019)Papamakarios, Sterratt, and Murray}]{Papamakarios2019-xp}
Papamakarios G, Sterratt D, Murray I (2019) Sequential neural likelihood: Fast likelihood-free inference with autoregressive flows. In: Proceedings of the Twenty-Second International Conference on Artificial Intelligence and Statistics, pp 837--848

\bibitem[{Ward et~al.(2022)Ward, Cannon, Beaumont, Fasiolo, and Schmon}]{Ward2022-fq}
Ward D, Cannon P, Beaumont M, et~al (2022) Robust neural posterior estimation and statistical model criticism. In: Advances in Neural Information Processing Systems, pp 33845--33859

\bibitem[{Tian et~al.(2025)Tian, Gao, Liu, Xu, Fujita, Aljuaid, and Wang}]{Tian2025-dd}
Tian WL, Gao P, Liu X, et~al (2025) Toward adaptive meta-gradient adversarial examples for visual tracking. IEEE Trans Reliab pp 1--14. \doi{10.1109/TR.2025.3569828}

\bibitem[{Xu et~al.(2025)Xu, Gao, Tang, Wang, and Yuan}]{Xu2025-xd}
Xu L, Gao P, Tang WJ, et~al (2025) Towards effective and efficient adversarial defense with diffusion models for robust visual tracking. Inf Fusion 124:103384. \doi{10.1016/j.inffus.2025.103384}

\bibitem[{Shahriari et~al.(2016)Shahriari, Swersky, Wang, Adams, and de~Freitas}]{Shahriari2016-fy}
Shahriari B, Swersky K, Wang Z, et~al (2016) Taking the human out of the loop: A review of bayesian optimization. Proc IEEE 104(1):148--175. \doi{10.1109/JPROC.2015.2494218}

\bibitem[{Nguyen(2019)}]{Nguyen2019-ww}
Nguyen V (2019) Bayesian optimization for accelerating hyper-parameter tuning. In: 2019 IEEE Second International Conference on Artificial Intelligence and Knowledge Engineering (AIKE), pp 302--305, \doi{10.1109/AIKE.2019.00060}

\bibitem[{Rumelhart et~al.(1986)Rumelhart, Hinton, and Williams}]{Rumelhart1986-jc}
Rumelhart DE, Hinton GE, Williams RJ (1986) Learning representations by back-propagating errors. Nature 323(6088):533--536. \doi{10.1038/323533a0}

\bibitem[{{Hecht-Nielsen}(1989)}]{Hecht1989-ox}
{Hecht-Nielsen} (1989) Theory of the backpropagation neural network. In: Proceedings of the International 1989 Joint Conference on Neural Networks, pp 593--605, \doi{10.1109/IJCNN.1989.118638}

\bibitem[{Tao et~al.(2024)Tao, Zheng, Wang, Qiu, and Stojanovic}]{Tao2024-ce}
Tao H, Zheng Y, Wang Y, et~al (2024) Enhanced feature extraction {YOLO} industrial small object detection algorithm based on receptive-field attention and multi-scale features. Meas Sci Technol 35(10):105023. \doi{10.1088/1361-6501/ad633d}

\bibitem[{Tao et~al.(2025)Tao, Huang, Wang, Qiu, and Vladimir}]{Tao2025-im}
Tao H, Huang Z, Wang Y, et~al (2025) Efficient feature fusion network for small objects detection of traffic signs based on cross-dimensional and dual-domain information. Meas Sci Technol 36(3):035004. \doi{10.1088/1361-6501/adb2ad}

\bibitem[{Sun et~al.(2025)Sun, Tao, and Stojanovic}]{Sun2025-dp}
Sun Y, Tao H, Stojanovic V (2025) End-to-end multi-scale residual network with parallel attention mechanism for fault diagnosis under noise and small samples. ISA Transactions 157:419--433. \doi{10.1016/j.isatra.2024.12.023}

\bibitem[{Li et~al.(2020)Li, Kroemer, Su, Veiga, Kaboli, and Ritter}]{Li2020-ia}
Li Q, Kroemer O, Su Z, et~al (2020) A review of tactile information: Perception and action through touch. IEEE Trans Rob 36(6):1619--1634. \doi{10.1109/TRO.2020.3003230}

\bibitem[{Kaboli et~al.(2016)Kaboli, Yao, and Cheng}]{Kaboli2016-ic}
Kaboli M, Yao K, Cheng G (2016) Tactile-based manipulation of deformable objects with dynamic center of mass. In: 2016 IEEE-RAS 16th International Conference on Humanoid Robots (Humanoids), pp 752--757, \doi{10.1109/HUMANOIDS.2016.7803358}

\bibitem[{Puentes et~al.(2024)Puentes, Morales, Pozo-Espin, and Moya}]{Puentes2024-ot}
Puentes K, Morales L, Pozo-Espin DF, et~al (2024) Enhancing control systems with neural network-based intelligent controllers. Emerg Sci J 8(4):1243--1261. \doi{10.28991/esj-2024-08-04-01}

\bibitem[{Huajun et~al.(2020)Huajun, Xinchi, Hang, and Shou}]{Huajun2020-yz}
Huajun Z, Xinchi T, Hang G, et~al (2020) The parameter identification of the autonomous underwater vehicle based on multi-innovation least squares identification algorithm. Int J Adv Robot Syst 17(2):1729881420921016. \doi{10.1177/1729881420921016}

\bibitem[{Rasul and Mukherjee(2024)}]{Rasul2024-jl}
Rasul T, Mukherjee K (2024) Data-driven approach for parameter estimation and control of an autonomous underwater vehicle. JEMS Maritime Sci 12(2):144--155. \doi{10.4274/jems.2024.10438}

\bibitem[{Vargas et~al.(2024)Vargas, Vasquez, De~la Barra, Charapaqui, Tapia-Yanayaco, Maldonado-Gómez, Mendoza-Arias, Altatorre, Ccellccaro, Bedoya-Castillo, Charapaqui, Nacarino, Rivera, Palomares, Ramirez-Chipana, Cornejo, Cornejo, and De~la Cruz-Vargas}]{Vargas2024-bl}
Vargas M, Vasquez Y, De~la Barra D, et~al (2024) Elbow-hand robotic exoskeletons for active and passive rehabilitation on post-stroke patients: A bioengineering review. HighTech Innov J 5(4):1170--1190. \doi{10.28991/hij-2024-05-04-020}

\bibitem[{Cornejo et~al.(2024)Cornejo, Cornejo, Vargas, Carvajal, Perales, Rodríguez, Macias, Canizares, Silva, Cubas, Jimenez, Lincango, Serrano, Palomares, Aspilcueta, Castillo-Larios, Evans, De~La Cruz-Vargas, Risk, Grossmann, and Elli}]{Cornejo2024-jb}
Cornejo J, Cornejo J, Vargas M, et~al (2024) {SY}-{MIS} project: Biomedical design of endo-robotic and laparoscopic training system for surgery on the earth and space. Emerg Sci J 8(2):372--393. \doi{10.28991/esj-2024-08-02-01}

\end{thebibliography}

\end{document}